%% file: arxiv.tex
\documentclass{fairmeta}
\usepackage{newpxtext,newpxmath}
\definecolor{hbblue}{HTML}{0668E1}
\colorlet{metablue}{hbblue}
\renewcommand{\title}[1]{\newcommand{\titlelist}{{\fontsize{23}{27}\selectfont\sffamily\bfseries\color{hbblue} #1}}}
\newcommand{\hbhl}[1]{\textcolor{hbblue}{\textbf{#1}}}

\input{math_commands.tex}

\usepackage{url}
\usepackage{enumitem}
\usepackage{amsmath}
\usepackage{makecell}
\usepackage{pifont}

\usepackage{wasysym}
\usepackage{xspace}
\usepackage{fvextra}
\usepackage{float}
\usepackage{colortbl}
\usepackage{listings}
\definecolor{forestgreen}{rgb}{0.13, 0.55, 0.13}
\definecolor{accessibleblue}{rgb}{0.00, 0.28, 0.67}
\definecolor{accessibleorange}{rgb}{0.80, 0.47, 0.00}
\definecolor{darkblue}{rgb}{0, 0, 0.5}
\definecolor{revisionpurple}{RGB}{128, 0, 128}
\newcommand{\cmark}{\textcolor{accessibleblue}{\ding{51}}}
\newcommand{\xmark}{\textcolor{accessibleorange}{\ding{55}}}
\newcommand{\mmark}{\textcolor{orange}{\LEFTcircle}}
\newcommand{\benchmarkname}{\hbhl{\textsc{HorizonBench}}\xspace}

\title{\benchmarkname: Long-Horizon Personalization with Evolving Preferences}

\author[1,2]{Shuyue Stella Li}
\author[2]{Bhargavi Paranjape}
\author[2]{Kerem Oktar}
\author[3]{Zhongyao Ma}
\author[2]{Gelin Zhou}
\author[2]{Lin Guan}
\author[2]{Na Zhang}
\author[2]{Sem Park}
\author[2]{Lin Chen}
\author[4]{Diyi Yang}
\author[1]{Yulia Tsvetkov}
\author[1,2]{Asli Celikyilmaz}

\affiliation[1]{University of Washington}
\affiliation[2]{Meta Superintelligence Labs}
\affiliation[3]{OpenAI}
\affiliation[4]{Stanford University}

\abstract{
Personal assistants now talk with the same user for months, and users' preferences change as their lives change; an assistant that keeps serving a preference the user has outgrown fails that user.
Real conversations do not record when or why a preference changed, and existing benchmarks either hold preferences fixed or have the user state the change, so they cannot tell whether a model forgot what the user said or failed to update its belief about the user.
We introduce \benchmarkname and its data generator, which samples how a user's preferences change and generates six months of conversation from that sequence; each change follows a life event that the user mentions without stating the new preference.
A \hbhl{human longitudinal preference dataset} that we collect and release, 83{,}024 ratings from 160 people over 13 weeks, sets how often and how much life events change preferences.
\benchmarkname\ contains 4{,}245 questions from 360 users, and when a preference has changed, its earlier value is one of the answer options.
All 25 frontier models we evaluate choose the outdated preference more often than evenly spread errors would predict, and the best reaches 52.8\% accuracy.
Two of the newest models score no higher when given only the relevant conversations than with the full history, yet one added sentence stating the change raises their accuracy by 11 to 22 points, and retrieval and memory methods do not recover the lost accuracy.
Remembering what a user said and knowing what the user prefers now are therefore \hbhl{separate capabilities}, and \benchmarkname, its generator, and the human dataset support research on both.
}

\date{\today}
\correspondence{\email{stelli@cs.washington.edu}}
\metadata[Code]{\url{https://github.com/stellalisy/HorizonBench}}
\metadata[Data]{\url{https://huggingface.co/datasets/stellalisy/HorizonBench}}

\begin{document}

\maketitle

\section{Introduction}\vspace{-3mm}

\begin{figure}[t]
    \centering
    \includegraphics[width=\linewidth]{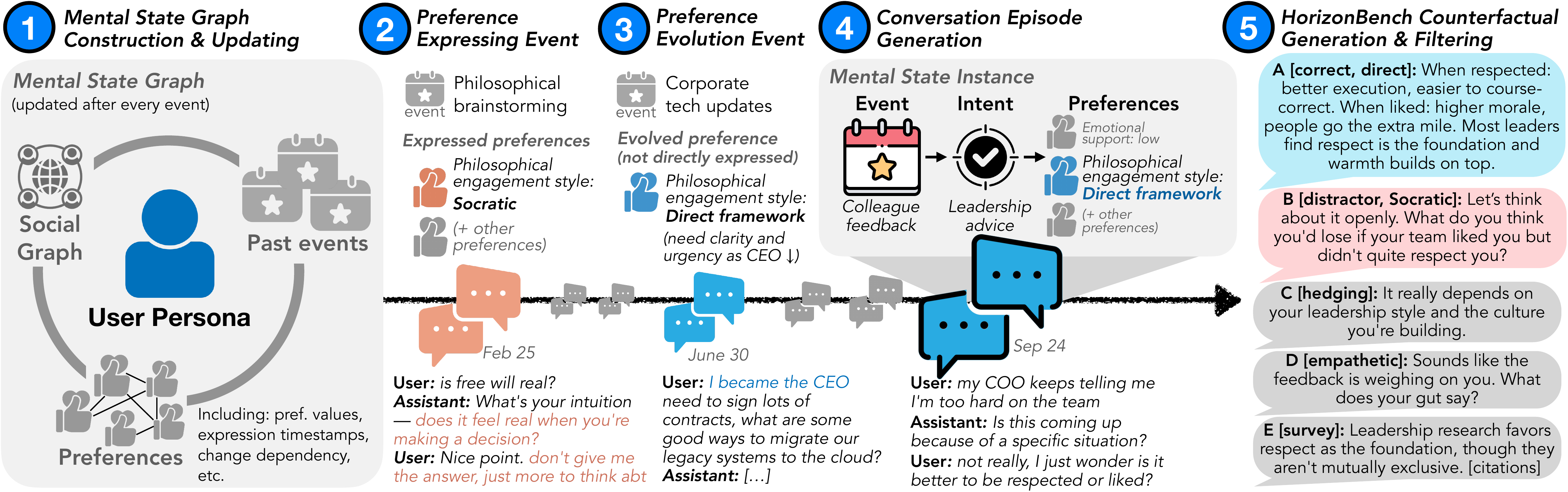}
    \caption{\benchmarkname pipeline overview. (1) The generator produces conversations from a structured mental state graph and records the preference state behind each conversation. (2) Preferences are established in conversations, then (3) changed by generated life events, which later conversations mention without stating the new values. (4) Later conversations are generated from the updated state and may express the new values. (5) The earlier values serve as pre-evolution distractors in \benchmarkname. See Appendix~\ref{app:example-task} for a complete benchmark item.
    }\vspace{-4mm}
    \label{fig:main}
\end{figure}

People increasingly interact with AI assistants over months of conversation~\citep{zhang2025riseaicompanionshumanchatbot}, and their preferences change during that time. Consider the user in Figure~\ref{fig:main}, who prefers Socratic philosophical discussion but, after becoming a chief executive, prefers direct frameworks without restating the preference. When she later asks whether it is better to be respected or liked, an assistant that answers in the Socratic style has retrieved her stated preference correctly but has not updated its belief about her. We call the task of identifying a user's current preference from a history in which later life events supersede earlier statements \hbhl{long-horizon personalization}. State variation can dominate stable trait differences in some user-behavior settings~\citep{harry2026fixedpsychologicalpersonasstate}, yet models and benchmarks commonly represent preferences as fixed traits.

Long-horizon personalization is difficult to measure, because natural conversations do not reveal a user's underlying preferences, so the time and cause of a change are unknown. Existing benchmarks test static preferences~\citep{zhao2025prefeval,li2025implexconv} or include changing information without an outcome that identifies reliance on a superseded preference~\citep{jiang2025personamemv2,tan2025membench,maharana2024locomo,wu2024longmemeval}, and synthetic user simulations do not record why preferences change~\citep{wang2024aipersona,chang2025chatbench}. Concurrent work models preferences that emerge across event-driven dialogues~\citep{liu2026perma}, repeated memory updates~\citep{uddin2026recall}, and implicit invalidation of earlier user states~\citep{chao2026stale}; Table~\ref{tab:comparison} compares these designs.

We introduce \benchmarkname\ and its data generator, the first resource that explicitly links a user's latent preference state to months of conversation and records the life event behind every preference change. The generator inverts the inference problem: it samples a sequence of user states from a structured mental state graph, in which typed dependency edges let one life event change several related preferences, and generates conversations from that sequence, so the preference behind every response and the event behind every change are known by construction. It also separates when a preference changes from when the user expresses it. A life event changes a preference without restating it, so the most recent explicit statement of a preference can be outdated when the question is asked, and the earlier value becomes a response option, the \hbhl{pre-evolution distractor}, whose selection identifies a failure to update. Together with a short-horizon condition and controlled rewrites of how explicitly preferences are expressed, this design separates \hbhl{failures to update a belief} about the user from \hbhl{failures to retrieve} what the user said. Research on preference construction and life-course change shows that preferences shift with experience and major life events~\citep{slovic1995construction,fleeson2001traits,specht2011stability}. To measure how likely a life event is to change preferences and by how much, we ran a \hbhl{13-week longitudinal study} in which 160 participants rated preferences related to a planned life event before the event and again two weeks later; after 17.5\% of completed or ongoing events, two to five of the six related preferences moved by at least three points on a seven-point scale, and the generator sets the probability and size of a preference change after a life event within these estimates (\S\ref{sec:calibration}).

\benchmarkname\ contains 4{,}245 five-option questions from 360 simulated users, each with a six-month history averaging ${\sim}$4{,}300 turns and ${\sim}$163K tokens (\S\ref{sec:benchmark}). A filter keeps only questions that five models answer incorrectly without the full history. All 25 frontier models we evaluate select the pre-evolution distractor more often than the 25\% expected under evenly spread errors, 23 of them significantly, and all 25 are less accurate on changed preferences than on unchanged ones. The best of the 25 models, Claude-opus-4.5, reaches 52.8\%, and ten models score below the 20\% random-guess rate. The pattern holds across context length, expression explicitness, and filter strictness, including when the preference was expressed 0--14 days before the question (\S\ref{sec:experiments}). Replacing the history with turns selected by BM25, dense retrieval, or recency lowers accuracy for every model, and Mem0 matches the full history without improving on it, because the conversation containing the life event never names the new preference (\S\ref{sec:memory}). Remembering what a user said and knowing what the user prefers now are therefore separate capabilities, and current models and memory methods lack the second.
We release the generator, the benchmark, the mental state graphs, and the human longitudinal dataset. Our contributions are:
\begin{enumerate}[itemsep=1pt,topsep=0pt,leftmargin=12pt]
  \item \textbf{Belief-update failure}, a failure mode in which a model answers with a user's superseded preference, observed in all 25 frontier models we evaluate (\S\ref{sec:experiments}).
  \item \textbf{\benchmarkname}, 4{,}245 questions with pre-evolution distractors and controlled variants of context length, expression explicitness, and filter strictness (\S\ref{sec:benchmark}).
  \item \textbf{A data generator} that constructs conversations from a structured mental state graph, records the life event behind every preference change, separates when a preference changes from when it is expressed, and calibrates the probability and magnitude of event-linked change against longitudinal human ratings (\S\ref{sec:generator}).
  \item \textbf{A human longitudinal preference dataset} of 83{,}024 ratings from 160 participants across seven surveys over 13 weeks, recording preferences tied to planned life events before and after each event, together with forecasts and written explanations of change (\S\ref{sec:calibration}, Appendix~\ref{app:human_calibration}).
\end{enumerate}

\begin{table}[t]
\centering
\renewcommand\arraystretch{1.05}
\caption{Comparison with related benchmarks. \textbf{Long Context}: 100K+ tokens. \textbf{Mental State Graph}: preferences with dependency edges. \textbf{Event-Driven Evo.}: preferences change through events. \textbf{Controlled Dims.}: difficulty factors varied independently. \textbf{Old-Value Distract.}: the previous value is a response option. \textbf{Human Calib.}: human data calibrate generation. \mmark\,=\,partial.}
\label{tab:comparison}\vspace{-3mm}
\small
\resizebox{\linewidth}{!}{
\begin{tabular}{lcccccc}
\toprule
\textbf{Method} & \makecell{\textbf{Long}\\\textbf{Context}} & \makecell{\textbf{Mental State}\\\textbf{Graph}} & \makecell{\textbf{Event-Driven}\\\textbf{Evo.}} & \makecell{\textbf{Controlled}\\\textbf{Dims.}} & \makecell{\textbf{Old-Value}\\\textbf{Distract.}} & \makecell{\textbf{Human}\\\textbf{Calib.}} \\
\midrule
\textsc{PrefEval \citep{zhao2025prefeval}} & \cmark & \xmark & \xmark & \xmark & \xmark & \xmark \\
\textsc{PersonaMem \citep{jiang2025personamem}} & \cmark & \xmark & \mmark & \xmark & \xmark & \xmark \\
\textsc{PersonaMem-v2 \citep{jiang2025personamemv2}} & \cmark & \xmark & \xmark & \xmark & \xmark & \xmark \\
\textsc{ImplexConv \citep{li2025implexconv}} & \xmark & \xmark & \xmark & \xmark & \xmark & \xmark \\
\textsc{LoCoMo \citep{maharana2024locomo}} & \mmark & \mmark & \xmark & \xmark & \xmark & \xmark \\
\textsc{MemBench \citep{tan2025membench}} & \mmark & \xmark & \xmark & \xmark & \xmark & \xmark \\
\textsc{PERMA \citep{liu2026perma}} & \cmark & \xmark & \cmark & \cmark & \xmark & \xmark \\
\textsc{Memora \citep{uddin2026recall}} & \mmark & \xmark & \mmark & \cmark & \xmark & \xmark \\
\textsc{STALE \citep{chao2026stale}} & \cmark & \mmark & \mmark & \cmark & \xmark & \xmark \\
\textbf{\benchmarkname} & \cmark & \cmark & \cmark & \cmark & \cmark & \cmark \\
\bottomrule
\end{tabular}
}\vspace{-2mm}
\end{table}

\section{Problem Formulation}\label{sec:problem}\vspace{-1mm}

\textbf{Latent user state.}\hspace{2mm}
A user's preferences are described by $K$ preference dimensions, each a typed attribute of a preference domain (e.g., \texttt{response\_structure} for emotional support), and dimension $k$ takes values in a finite set $\mathcal{V}_k$. The user's latent state at conversation $t \in \{1, \ldots, T\}$ is $u_t = (v_{t,1}, \ldots, v_{t,K})$. The state is piecewise constant: a life event $\ell_t$ may update a subset $S_t \subseteq \{1, \ldots, K\}$ of dimensions, $u_t = f_{\text{evo}}(u_{t-1}, \ell_t)$, and otherwise $u_t = u_{t-1}$. Each conversation $C_t$ is generated from $u_t$, and a model observes only the history $H_t = (C_1, \ldots, C_t)$, never $u_t$ or which dimensions changed.

\textbf{User-conditioned reward.}\hspace{2mm}
Given a conversational context $x$ in conversation $t$, the user's reward for an assistant response $r$ is $R(x, r; u_t)$, which measures how well $r$ fits the user's current state. The true reward depends on the history only through $u_t$. A model acting as a reward model assigns a score $s(r \mid H_t, x)$ without access to $u_t$, so it must infer the relevant part of $u_t$ from $H_t$. When a dimension has changed, a score that still conditions on its earlier value ranks responses by a reward the user no longer holds.

\textbf{Evaluation.}\hspace{2mm}
Each item queries one dimension $k$ at a test turn with context $x$ and offers five candidate responses. The correct response $r^*$ realizes the current value $v^* = v_{t,k}$, and four alternatives $r_1, \ldots, r_4$ are written to differ from it only in realizing other values $v_1, \ldots, v_4 \in \mathcal{V}_k$, so $R(x, \cdot\,; u_t)$ ranks $r^*$ highest. A model is correct when it does the same, $s(r^* \mid H_t, x) > s(r_j \mid H_t, x)$ for all $j$; for a generative model that selects a response directly, the selection is the argmax of its implicit scores. This is the listwise form of reward-model evaluation~\citep{lambert2024rewardbench}, conditioned on a user state the model must infer.
An item is \emph{evolved} if the preference has changed since the user first expressed it, and \emph{static} otherwise. On evolved items whose earlier value was expressed in the history, one alternative realizes that value, $v_{\text{old}}$; we call it the \emph{pre-evolution distractor}.
Besides accuracy, we report two measures of belief-update failure. The \emph{distractor rate} is the share of wrong answers that select the pre-evolution distractor; if errors were spread evenly over the four wrong options, it would be 25\%, the uniform-error reference. The \emph{evolved--static gap} is accuracy on evolved items minus accuracy on static items.

\textbf{Generating histories from states.}\hspace{2mm}
Because $u_t$ is never observed in natural conversation, the same history can be consistent with several sequences of user states, so neither $R$ nor the correct response can be established from real data. We invert this ``inferring state from history'' problem in the data generator (\S\ref{sec:generator}), which samples a sequence of user states $u_{1:T}$ and produces $H_T$ from it. Therefore, the reward labels of every item and the life event behind every change are known by construction, and each change has one designated current value.

\section{\benchmarkname Data Generator}\label{sec:generator}

The generator models both the user and the assistant as stateful agents. Its user state augments the preference state $u_t$ of \S\ref{sec:problem} with a persona profile, and the assistant state $a_t$ likewise comprises a persona profile and typed preference attributes. The generator's central data structure is a \emph{mental state graph} $\mathcal{G}$ that maintains each agent's persona, social connections, life events, and preferences. Preferences are linked by typed dependency edges (e.g., dietary preferences $\to$ restaurant choices $\to$ social outing plans), and life events alter preference values with changes propagating through these edges. The generator records the life event behind every change and tracks when each preference changes separately from when it is expressed. A preference that has not been expressed for a set number of days becomes \emph{stale}, and the generator can then \emph{recall} it by placing a later turn whose response must reflect the preference's current value; each such turn is a candidate benchmark question (\S\ref{sec:temporal_recall}).
The generator conditions every call on $\mathcal{G}$, which keeps each agent consistent across six months of conversation.
 
\subsection{Event Sampling}
 
Each conversation $C_t$ is organized around an episode event $e_t$, the occasion for the interaction (e.g., rehearsing for a debate), which is distinct from the life events $\ell_t$ that change preferences (\S\ref{sec:calibration}). Episode events are sampled conditional on both agent states and event history, $e_t \sim p(e \mid u_{t-1}, a_{t-1}, e_{1:t-1};\, \boldsymbol{\beta})$, preventing unrealistic repetition from independent draws. Category weights $\boldsymbol{\beta}$ follow empirical AI-chatbot interaction distributions~\citep{zhang2025riseaicompanionshumanchatbot} (emotional support 26\%, collaborative storytelling 26\%, romantic interactions 22\%). When a stale preference is selected for temporal recall, the sampled event is constrained to that preference domain, ensuring the recall turn arises in natural conversational context.
 
\subsection{Preference Evolution}
\label{sec:calibration}
 
Preferences do not change in isolation. A career change may jointly affect work preferences, commute habits, and financial priorities. Typed dependency edges in $\mathcal{G}$ represent such relationships, $u_t = f_{\text{evo}}(u_{t-1}, \ell_t, \mathcal{G};\, \boldsymbol{\phi}_{\text{evo}})$, where $\boldsymbol{\phi}_{\text{evo}}$ collects the evolution parameters described below. Each life event changes the user's preferences with probability $p_{\text{evo}} = 0.15$ and otherwise leaves $u_t$ unchanged. When a change occurs, a connected set of 2--5 preferences is selected, and an LLM jointly constructs the event and a qualitatively distinct new value for each preference, conditioned on the user's current state, and the generator records the event with the change. A short conversation then mentions the event without stating the new values.

\textbf{Human longitudinal preference dataset.}\hspace{2mm}
Setting how often a life event changes a person's preferences, and by how much, requires measuring the same people's preferences before and after events in their lives. Studies of life-course change follow broad personality traits across years~\citep{specht2011stability} and do not record the everyday preferences an assistant serves around specific events, so we collect a longitudinal dataset and release it with the benchmark (Appendix~\ref{app:human_calibration}). The study tracks 52 everyday preferences in 13 categories: six of concrete preferences, such as food, activity, and daily rhythms, and seven of value-based preferences, such as risk, planning, and work style. The items draw on economic theories of preference~\citep{lancaster1966new,falk2018global} and on personality psychology~\citep{becker2012relationship} (Appendix~\ref{app:pref-grounding}). Over 13 weeks, 160 participants completed seven surveys about two weeks apart. At every survey, they rated the 52 preferences on a seven-point scale and predicted their ratings at each later survey. They also described a new experience they had planned for the next two weeks and listed and rated six of their preferences that it might affect. At the next survey, they reported whether the planned event had happened and rated the same six preferences again. The dataset contains 83{,}024 ratings, 960 reports of whether a planned event happened, participants' written explanations of change, and a final survey on perceived change and preference-aware AI (Table~\ref{tab:human_dataset_contents}). Beyond calibrating the generator, it supports research on preference stability, people's forecasts of their own preferences, and change around life events.

\textbf{Calibration.}\hspace{2mm}
We use the dataset to set two parameters of the generator: the probability $p_{\text{evo}}$ that a life event changes the user's preferences, and the size of each change. A change in the generator moves two to five related preferences to clearly different values, and we measure how often the same pattern occurs in people. For each planned event that happened, we compare the six preferences the participant tied to it, rated before the event and again two weeks later, and count the event as changing preferences when two to five of the six ratings moved by at least three points on the seven-point scale. Of the 470 events that happened or were underway and had all six ratings at both surveys, 17.5\% changed preferences in this sense (95\% CI 13.2--21.9\%, resampling participants), and the ratings that moved changed by 3.64 points on average (95\% CI 3.51--3.77). We set $p_{\text{evo}}=0.15$, within this interval, and each change moves a preference to a qualitatively different value, matching the large size of the observed changes (Table~\ref{tab:human_calibration}).

\begin{table}[t]
\centering
\small
\caption{Human estimates (95\% CIs) and the corresponding generator settings.}\vspace{-3mm}
\label{tab:human_calibration}
\resizebox{\linewidth}{!}{
\begin{tabular}{llll}
\toprule
\textbf{Quantity} & \textbf{Human estimate} & \textbf{Generator setting} & \textbf{Definition in human data} \\
\midrule
Share of events that change preferences & 0.175 [0.132, 0.219] & $p_{\text{evo}}=0.15$ & 2--5 of 6 ratings move $\geq 3$ points \\
Size of a change & 3.64 [3.51, 3.77] points & Qualitatively different value & Mean move of those ratings \\
\bottomrule
\end{tabular}}\vspace{-3mm}
\end{table}
 
The evolution event itself does \emph{not} refresh a preference's expression timestamp. A preference updated on day 45 remains stale from its previous expression on day 10 and may become immediately eligible for recall. Later conversations can express the updated value and refresh its timestamp, so evolved items vary in how much post-evolution evidence precedes the test. This design keeps the pre-evolution value available as a response option and lets us measure how expression frequency affects model choices (\S\ref{sec:frequency}).
 
\subsection{Conversation Generation}\vspace{-1mm}
 
An \emph{outline} is generated first, specifying which preferences appear, in what order, and with what expression style, enforcing structural constraints such as placing temporal recall turns in contextually appropriate positions. Content is then generated turn by turn conditional on the outline, $m_{t,n}^u \sim p_u(m \mid u_t, e_t, H_{t,n-1})$ and $m_{t,n}^a \sim p_a(m \mid a_t, m_{t,n}^u, H_{t,n-1})$, where $m_{t,n}^u$ and $m_{t,n}^a$ are the user and assistant messages of turn $n$ in $C_t$, and $H_{t,n-1}$ is $H_{t-1}$ followed by the first $n-1$ turns of $C_t$. The assistant persona is generated per-event to match the interaction context (e.g., a career counselor for a job change).
To control how explicitly preferences are expressed, we produce variants of the original (\emph{neutral}) conversations by rewriting only preference-expressing user turns and the immediately following assistant turns. In the \emph{explicit} variant, preferences are stated with clear, varied phrasing while preserving the original register and structure. In the \emph{implicit} variant, preferences are conveyed only through tangential remarks and contextual cues. Because the conversation content, preference structure, sequence of preference changes, and benchmark items remain identical across variants, only the expression of preferences varies (\S\ref{sec:controlled}).
 
\subsection{Expression Tracking and Temporal Recall}\label{sec:temporal_recall}\vspace{-1mm}
 
Tracking when each preference was expressed lets the benchmark tell apart a model that never saw the preference, one that retrieved it incorrectly, and one that failed to update it. Conversations are placed on a six-month calendar, with $d_t$ the day of conversation $t$. Each preference $i$ maintains the day $\tau_i$ of its last conversational expression, refreshed after each conversation, and preferences whose staleness $d_t - \tau_i$ exceeds a threshold (default 30 days) become eligible for temporal recall, sampled proportionally to staleness and optionally weighted toward evolved preferences.
 
Every recalled preference has been expressed in the complete benchmark history. When the preference has evolved, its originally expressed value is injected as a distractor (\S\ref{sec:counterfactual}), and the associated life event appears in a dedicated conversation; some histories also contain later expressions of the updated value.
All generator parameters (timeline length, expression explicitness, evolution probability, staleness threshold, user demographics) are configurable.

\section{\benchmarkname Benchmark}\label{sec:benchmark}\vspace{-1mm}

\benchmarkname is one instantiation of the data generator. Each benchmark item is a 5-option multiple-choice question embedded within a conversation history. Given the full history, the model must identify which assistant response best aligns with the user's \emph{current} preferences, the response ranked highest by the user-conditioned reward $R(x, \cdot\,; u_t)$ of \S\ref{sec:problem}.
 
\subsection{Counterfactual Evaluation and History Filtering}\label{sec:counterfactual}
 
For each temporal recall turn, one attribute $k$ of the recalled preference is sampled uniformly, and its current value $v^* = v_{t,k}$ defines the correct response. An LLM proposes four maximally contrasting alternative values $\{v_1, \ldots, v_4\}$ for $k$. If the value of $k$ has changed since the user first expressed it, the originally expressed value $v_{\text{old}}$ replaces one of them as the pre-evolution distractor. For each alternative $v_j$, an LLM rewrites the assistant turn into a contextually appropriate response $r_j$ that fits $v_j$.
 
Items are then passed through a \textbf{5-LLM consensus history filter}. Each candidate is presented to five validation models (GPT-4.1, Gemini-3-flash, o4-mini, Claude-haiku-4.5, Claude-opus-4.6) \emph{without} the full conversation history and passes only if all five answer incorrectly. This removes items that the validators can answer without the conversation history. The filter threshold is tunable. Strict consensus (5/5) retains the hardest items, while majority ($\geq$3/5) admits items that are easier to answer without the history, so the threshold sets item difficulty (\S\ref{sec:controlled}).
 
\subsection{Dataset Statistics}\label{sec:dataset}
 
\benchmarkname is produced by three frontier generators (Claude-sonnet-4.5 with 1{,}052 items from 120 users, o3 with 981 items from 120 users, and Gemini-3-flash with 2{,}212 items from 120 users), which allows results to be compared across generators (\S\ref{sec:validation}). The pipeline generates 57{,}390 conversations totaling 1.6M turns across 360 users, with six-month timelines averaging 142--168 conversations and 3{,}600--5{,}000 turns per user across 30 preference domains. Of 45{,}006 candidates, 4{,}245 pass the strict 5-LLM filter (9.4\% pass rate), of which 59\% involve evolved preferences and 41\% static. Appendix~\ref{app:example-task} shows a complete item.

\subsection{Benchmark Validation}\label{sec:validation}\vspace{-1mm}
 
\textbf{Results are consistent across generators.}\hspace{2mm}
Claude models score $-4.8$\,pp on Claude-sonnet-4.5-generated items relative to the other generators, and OpenAI models $-2.6$\,pp on o3-generated items. 
 
\textbf{Items are answerable.}\hspace{2mm}
An item's answer is correct when the new value follows from the life event and the correct option expresses that value; we test each condition directly. Given the earlier value and the user's turns in the life-event conversation, the majority of three annotators chose the designated value among the item's five values on 86.7\% of 60 items (all three on 70.0\%, Fleiss $\kappa = 0.79$). Reading one option alone with no history, gpt-6-astra and gemini-4-argon named its recorded value for 81.6\% and 82.0\% of 5{,}000 options. Details and results on the items all three annotators answered correctly are in Appendix~\ref{app:answerability}.

\textbf{Response quality checks.}\hspace{2mm}
For 4{,}235 items with three annotations each, individual positive ratings were 99.7\% for situational fit, 99.7\% for conversation-flow realism, and 99.4\% for assistant-persona consistency; every item received a positive majority rating on all three dimensions (Appendix~\ref{app:human_annotation}).

\section{Experiments}\label{sec:experiments}\vspace{-2mm}

We use \benchmarkname\ to answer three questions:
\begin{enumerate}[itemsep=0pt,topsep=2pt,leftmargin=30pt]
  \item[\textbf{RQ1.}] Do models update what they know about a user when a life event changes a preference?
  \item[\textbf{RQ2.}] How does the failure change with context length, expression frequency, expression explicitness, and item difficulty?
  \item[\textbf{RQ3.}] Do retrieval and memory methods recover the lost accuracy?
\end{enumerate}

\subsection{Experimental Setup}\label{sec:setup}\vspace{-1mm}
We evaluate 25 models spanning three families, including 8 OpenAI models (gpt-4o, gpt-4.1, gpt-5, gpt-5-mini, o1, o3, o3-mini, o4-mini), 9 Gemini models (2.0-flash-lite through 3.1-pro), and 8 Claude models (3.5-haiku through opus-4.5). Each model receives the full conversation history spanning six months (${\sim}$4{,}300 turns, averaging ${\sim}$163K tokens per prompt with some exceeding 250K) and selects which of five assistant responses best aligns with the user's demonstrated preferences. We report results on the combined multi-generator dataset (4{,}245 items from 360 users across three generators). Items exceeding a model's context window are truncated by removing earlier conversation turns. The short-horizon condition (\S\ref{sec:controlled}) provides a complementary test on shorter histories. Sign and Wilcoxon tests are computed over the 25 per-model values; we also report user-clustered bootstrap intervals over the 346 users with items. All confidence intervals are 95\% percentile bootstrap CIs with $B{=}10{,}000$ resamples.

\begin{figure}[t]\vspace{-2mm}
    \centering
    \includegraphics[width=0.86\linewidth,height=4.7cm]{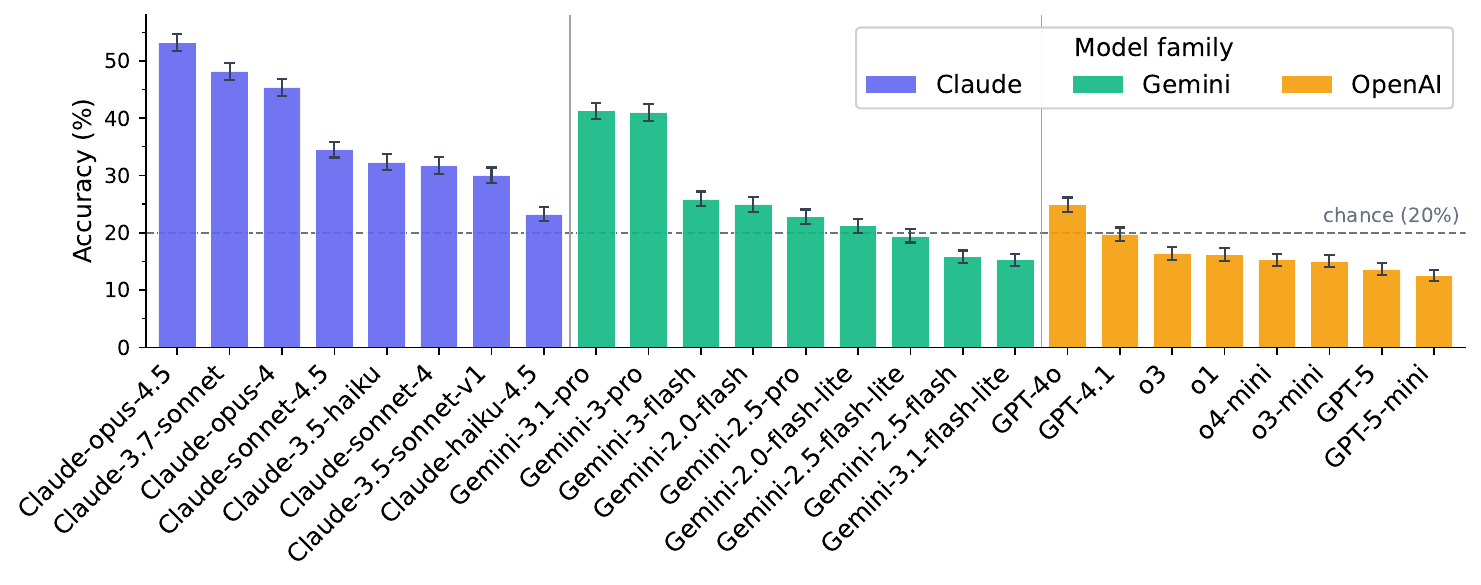}\vspace{-4.5mm}
    \caption{Per-model accuracy on \benchmarkname (4{,}245 items). Bars are colored by model family with bootstrap 95\% CIs ($B{=}10{,}000$). Dashed line marks the 20\% random-guess rate.}
    \label{fig:accuracy}\vspace{-3mm}
\end{figure}

\subsection{RQ1: Models Answer with Superseded Preferences}\label{sec:main_results}\label{sec:preevo}\vspace{-1mm}

Figure~\ref{fig:accuracy} reports per-model accuracy on the 4{,}245 items (full results in Table~\ref{tab:full_results}). The best model, Claude-opus-4.5, achieves 52.8\%. Ten of 25 models score below the 20\% random-guess rate, including seven of eight OpenAI models and three Gemini models. Position-debiased accuracy is within 1\,pp of raw accuracy for every model. Because the filter keeps only items that five validators answer incorrectly without the history, which removes items whose response options alone point to the correct answer, accuracy can fall below the 20\% random-guess rate, on static items as well (gpt-5-mini 13.3\%, gpt-5 15.4\%). The distractor rate and the evolved--static gap therefore measure the update failure relative to each model's own accuracy.

We additionally evaluate five open-weight models from the Llama, OLMo, and Qwen families on 1{,}011 items from the Claude-sonnet-4.5-generated subset, with left truncation for items exceeding their context windows (Table~\ref{tab:openweight_results}). All five score below 20\% (11.6--17.1\%) and have lower accuracy on evolved than static items, so the evolved--static gap also appears in open-weight models.

\begin{minipage}{\textwidth}\vspace{3mm}
    \begin{minipage}{0.49\textwidth}
    \centering
    \includegraphics[width=\linewidth]{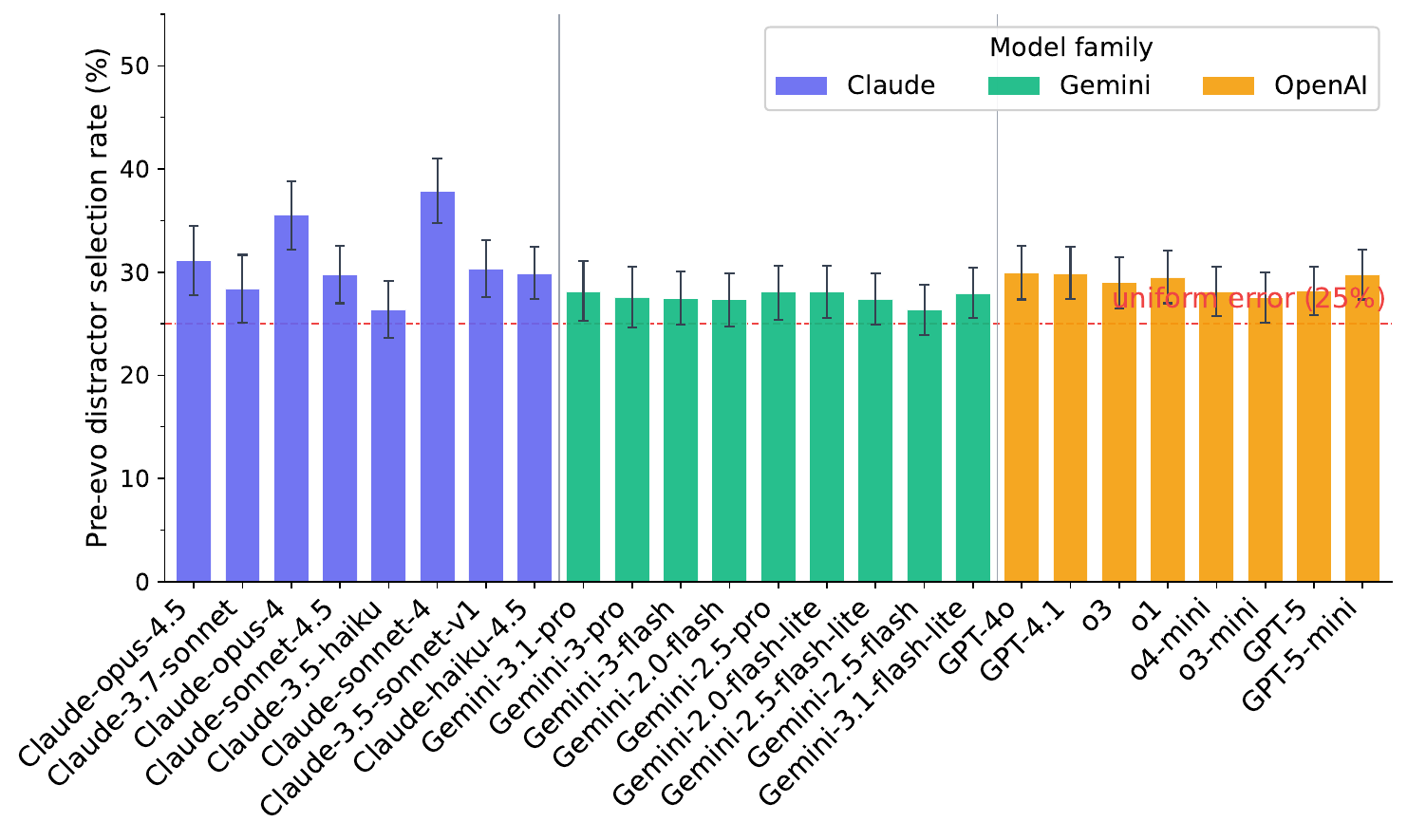}\vspace{-3mm}
    \captionof{figure}{Distractor rate by model. The dashed line marks the 25\% uniform-error reference (1 of 4 wrong options); all 25 models exceed it.
    }
    \label{fig:distractor_rate}
    \end{minipage}
\hfill
    \begin{minipage}{0.49\textwidth}
    \centering
    \includegraphics[width=\linewidth]{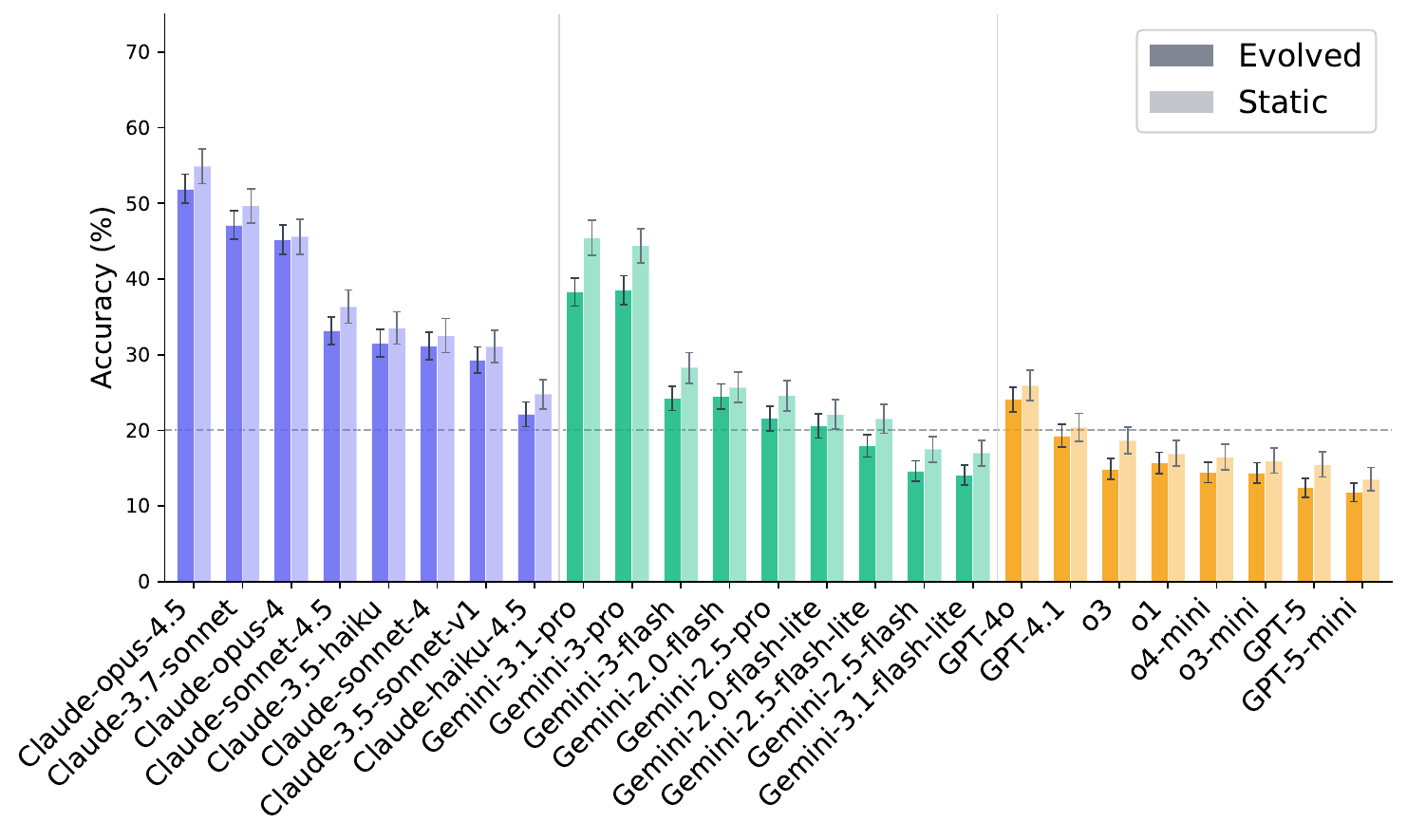}\vspace{-3mm}
    \captionof{figure}{Accuracy on evolved (dark) vs.\ static (light) preferences with 95\% bootstrap CIs on the combined dataset (4{,}245 items). All 25 models perform worse on evolved items (sign test $p < 0.001$).}
    \label{fig:evo_static}
\end{minipage}
\end{minipage}\vspace{3mm}

\textbf{Distractor rate.}\hspace{2mm}
All 25 models exceed the 25\% uniform-error reference (Figure~\ref{fig:distractor_rate}), with a mean distractor rate of 29.2\% (user-clustered 95\% CI 26.9--31.4\%) and a maximum of 37.8\%. Twenty-three of the 25 are individually significant under a one-sided binomial test ($p<0.05$). Claude-opus-4.5, the most accurate model, selects it on 31.1\% of its wrong answers. Within the OpenAI family, reasoning and non-reasoning models show similar mean distractor rates (28.7\% vs.\ 29.9\%), so inference-time reasoning alone does not resolve the failure.

\textbf{Evolved--static gap.}\hspace{2mm}
All 25 models perform worse on evolved preferences than static ones ($-$2.7\,pp average, user-clustered 95\% CI $-$4.2 to $-$1.1, sign test 25/25 negative, $p < 0.001$, Figure~\ref{fig:evo_static}). The gap is largest for Gemini models ($-$3.5\,pp) and smallest for OpenAI ($-$2.1\,pp). Static accuracy also stays low, so retrieval is a separate source of difficulty.

\subsection{RQ2: The Failure Persists across Controlled Conditions}\label{sec:controlled}\vspace{-1mm}

The generator's controls let us vary context length, expression frequency, expression explicitness, and item difficulty one at a time while holding the preference changes fixed. Table~\ref{tab:controlled} reports the evolved--static gap and the distractor rate under each condition. The neutral, explicit, and implicit conditions use the 981 o3-generated items, whose distractor rate (38.2\%) is the highest of the three generators (29.2\% on the full benchmark; Table~\ref{tab:by_generator}).

\begin{table}[t]
    \centering
    \caption{Belief-update metrics across controlled conditions, averaged over the 25 models. Neutral, explicit, and implicit rows use the 981 o3-generated items; the short-horizon and relaxed-filter rows use their own item sets (per-model results in Appendix~\ref{app:controlled_detail}). \emph{Evo$\Delta$}: evolved--static gap. \emph{Dist.\%}: distractor rate (uniform-error reference 25\%). The two count columns give how many models show each pattern.}
    \label{tab:controlled}\vspace{-2mm}
    \resizebox{0.8\linewidth}{!}{
    \begin{tabular}{l r@{\hskip 1mm}l r@{\hskip 1mm}l @{\hskip 0mm}r@{\hskip 1mm}l r r r r}
    \toprule
    \textbf{Condition} & \textbf{Overall} && \textbf{Static} && \textbf{Evolved} && \textbf{Evo$\boldsymbol{\Delta}$} & \textbf{Evo$\boldsymbol{\Delta}$${<}$0} & \textbf{Dist.\%} & \textbf{Dist.${>}$25\%} \\
    \midrule
    Neutral (unmodified) & 27.0 && 28.9 && 26.0 && $-$2.9 & 22/25 & 38.2 & 25/25 \\
    Short horizon       & 30.6 &$\uparrow$ & 31.4 &$\uparrow$ & 25.7 &$\downarrow$ & $-$5.7 & 25/25 & 47.7 & 25/25 \\
    Explicit rewrite    & 27.9 &$\uparrow$ & 29.0 &$\uparrow$ & 27.2 &$\uparrow$ & $-$1.8 & 16/25 & 49.7 & 25/25 \\
    Implicit rewrite    & 26.4 &$\downarrow$ & 28.6 &$\downarrow$ & 25.0 &$\downarrow$ & $-$3.6 & 20/25 & 58.8 & 25/25 \\
    Relaxed filter      & 38.9 &$\uparrow$ & 40.3 &$\uparrow$ & 37.9 &$\uparrow$ & $-$2.5 & 22/25 & 54.3 & 25/25 \\
    \bottomrule
    \end{tabular}
    }
\end{table}

\textbf{Context length.}\hspace{2mm}
Because the generator tracks when each preference was last expressed, we can regenerate histories with the staleness threshold (\S\ref{sec:temporal_recall}) relaxed, so a preference is tested 0--14 days after its last expression instead of at least 30 days later (${\sim}$95K tokens per history, 120 users, 1{,}778 items). Static accuracy rises slightly (31.4\% vs.\ 28.9\%), consistent with an easier retrieval task in shorter context, yet the gap widens to $-$5.7\,pp (25/25 models negative, Wilcoxon $p < 10^{-7}$) and the distractor rate rises to 47.7\% (25/25 above the uniform-error reference, Wilcoxon $p < 10^{-7}$).

\textbf{Expression frequency.}\label{sec:frequency}\hspace{2mm}
The underlying state graph of \benchmarkname\ also allows us to investigate the effect of preference expression frequency on the model's final choice. We count the conversations expressing the pre- and post-evolution values in each model-visible history. The new value is expressed more often in most histories (65.7--69.2\% of expressions of either value, and the more frequent one in 65.6--70.5\% of matched histories; Table~\ref{tab:frequency_analysis}), yet model accuracy does not follow it. Averaged over the 25 models, accuracy when the new value is more frequent is 21.5\% for Claude-sonnet-4.5-generated histories and 20.8\% for o3-generated ones, against 29.2\% and 34.4\% when the old value is more frequent; for Gemini-3-flash it is 25.5\% against 22.3\%. Model choices are largely independent of how often each value is expressed.

\textbf{Expression explicitness.}\hspace{2mm}
Prior work identifies implicit preference expression as a main source of difficulty~\citep{li2025implexconv,jiang2025personamemv2}. Starting from the neutral conversations, we evaluate the explicit and implicit rewrites of \S\ref{sec:generator}, which follow the methodology of~\citet{li2025implexconv}. Overall accuracy decreases monotonically (27.9\% explicit, 27.0\% neutral, 26.4\% implicit). The evolved--static gap widens from $-$1.8\,pp (explicit) to $-$2.9\,pp (neutral) to $-$3.6\,pp (implicit), and the distractor rate rises from 49.7\% (explicit) to 58.8\% (implicit). Both rewrites raise the distractor rate above the neutral 38.2\%: explicit wording makes the earlier value as salient as the new one, and implicit wording leaves the new value harder to infer even though it is expressed more often (\S\ref{sec:frequency}).

\textbf{Filter strictness.}\hspace{2mm}
Relaxing the 5-LLM filter (\S\ref{sec:counterfactual}) from strict consensus (5/5) to majority ($\geq$3/5) raises average accuracy by $+$11.9\,pp (every model positive, paired Wilcoxon $p < 10^{-7}$). The evolved--static gap is $-$2.5\,pp (22/25 models negative, Wilcoxon $p < 10^{-4}$), and every model has a distractor rate above the uniform-error reference (54.3\%, 25/25). The filter sets item difficulty, and the belief-update pattern holds at both thresholds.

\textbf{Matched evidence.}\hspace{2mm}
For 200 items, we use the generator's records to keep only the conversations expressing the earlier value, the life-event conversation, and later expressions of the new value (median 5.3K tokens), with the same test conversation, question, and options (Appendix~\ref{app:matched}). gpt-6-astra and gemini-4-argon score 27.0\% and 58.0\%, no higher than with the full history (28.5\% and 63.5\%). One added user sentence stating the change raises accuracy by 22.0 [14.5, 29.5] and 11.0 [3.0, 19.0] points, and a paraphrase that never names the value is within 3.6 points of it. With every relevant conversation in context, the models fail to infer the change.

\subsection{RQ3: Retrieval Cannot Surface the Evidence of a Change}\label{sec:memory}\vspace{-1mm}

Long-horizon personalization is commonly treated as a memory problem, so we replace the full history with a smaller context while holding fixed the model that answers the question (the reader) and the scoring. BM25~\citep{robertson2009bm25}, dense retrieval with a locally run 384-dimensional encoder, and dense retrieval with \texttt{text-embedding-3-small} each return the ten highest-scoring turns in timestamp order, following \citet{wu2024longmemeval}, and a recency baseline returns the ten most recent turns. All four run on the 2{,}484 evolved items.

\begin{table}[t]
\centering
\small
\caption{Accuracy (\%) on the 2{,}484 evolved items when a retrieval method replaces the full history.}
\label{tab:memory}\vspace{-2mm}
\resizebox{0.7\linewidth}{!}{
\begin{tabular}{lrrrrr}
\toprule
\textbf{Model} & \textbf{Full history} & \textbf{BM25} & \textbf{Recency} & \textbf{Dense 384d} & \textbf{Dense 1536d} \\
\midrule
Claude-opus-4.5   & 52.1 & 42.2 & 42.8 & 40.3 & 40.6 \\
Claude-sonnet-4.5 & 33.3 & 27.5 & 26.4 & 24.8 & 25.6 \\
GPT-4o            & 24.1 & 17.5 & 15.5 & 17.8 & 18.0 \\
Claude-haiku-4.5  & 20.5 & 16.3 & 13.0 & 15.2 & 15.3 \\
GPT-4.1           & 19.3 & 18.0 & 14.1 & 17.6 & 17.5 \\
\bottomrule
\end{tabular}
}\vspace{-1mm}
\end{table}

Every retrieval condition falls below the full history for every model, and the strongest model loses the most: Claude-opus-4.5 drops 9.9\,pp under BM25 and up to 11.8\,pp under dense retrieval (Table~\ref{tab:memory}). The distractor rate reaches 37--47\%, against 29.7--31.1\% for the same readers with the full history. Mem0~\citep{chhikara2025mem0}, which builds memories with a language model as conversations arrive and never sees the question, reaches 19.9\% with \texttt{gpt-4.1} as the reader (1{,}492 items with a pre-evolution distractor), against 19.3\% for the same reader on the full history. A note-taking agent built on gpt-6-astra and gemini-4-argon, which writes notes from the matched-evidence conversations with an instruction to record changed preferences and then answers from them, scores 21.5\% and 42.0\%, below reading those conversations (27.0\% and 58.0\%; Appendix~\ref{app:matched}).

The generator's records explain this. Because the generator stores the conversation in which each life event changed a preference, retrieval can be scored against it: BM25 retrieves that conversation on 18.5\% of items, against 22.1\% for a random draw of the same size, and a statement of the current value on 35.9\%, against 31.4\%. The two dense encoders select largely different turns (21.1\% mean Jaccard overlap) yet differ by at most 0.8\,pp, so which turns reach the model barely matters. An event turn describes a career change or a diagnosis without naming a preference value, so similarity to a question about that preference does not surface it, and no selection from the history can supply evidence that the history states only implicitly.

\section{Related Work}
\vspace{-1mm}

\textbf{Personalization and long-term memory benchmarks.}\hspace{2mm}
Existing benchmarks test whether models retrieve stated preferences and facts from long histories~\citep{zhao2025prefeval,jiang2025personamem,maharana2024locomo,tan2025membench,wu2024longmemeval,liu2024lost}, often treating implicit expression as the main source of difficulty~\citep{jiang2025personamemv2,li2025implexconv,guo2026realpref}. Those that include changing information have no outcome that identifies continued reliance on a superseded preference, so a retrieval failure and an update failure produce the same error. Concurrent work models evolving user state through emerging preferences~\citep{liu2026perma}, repeated memory updates~\citep{uddin2026recall}, and implicit state invalidation~\citep{chao2026stale}. ICF-Bench tests whether models follow an explicit update that reverses an earlier preference~\citep{qian2026icf}, and memory agents learn to update stored notes~\citep{xu2026memcoe} or search raw chat logs repeatedly~\citep{li2026refind}. \benchmarkname\ differs in three ways: the earlier value appears among the response options, so selecting it measures the update failure directly; every change is tied to a recorded life event; and context length and expression vary while the transition is held fixed (Table~\ref{tab:comparison}). The resulting error pattern connects to findings that models favor prior evidence over newer context~\citep{xu2024knowledge}.

\vspace{-0.5mm}
\textbf{Preference change.}\hspace{2mm}
Psychological research shows that preferences are partly constructed in context and shift with experience, roles, and life events~\citep{slovic1995construction,schwarz2007attitude,fleeson2001traits,elder1998lifecourse,hitlin2004values,specht2011stability}, which motivates the generator's event-linked evolution; our human longitudinal dataset calibrates its parameters and is released for direct study of preference change (\S\ref{sec:calibration}). Temporal recommender systems model preference drift from timestamped behavior~\citep{koren2009collaborative,rafailidis2015modeling,rabiu2020recommender}. \benchmarkname\ targets the conversational case, in which an event implies a new preference before the user states it.

\vspace{-0.5mm}
\textbf{User simulation.}\hspace{2mm}
The generator builds on LLM agent simulation~\citep{park2023generative} and synthetic life-long personalization data~\citep{wang2024aipersona,chang2025chatbench}, which lack structured preference transitions with known causes. Structured models of how preferences co-vary support inferring a user's preferences from few interactions~\citep{bose2026pep}, and evidence that state variation can dominate stable trait differences in some user-behavior settings~\citep{harry2026fixedpsychologicalpersonasstate} motivates modeling preferences as changing states. Appendix~\ref{app:related} extends this discussion.

\section{Conclusion}\vspace{-1mm}

We introduced \benchmarkname\ and its generator, the first resource that explicitly links a user's latent state to months of conversation and records the cause of every preference change, which provides supervision that natural conversation data cannot. All 25 frontier models choose the user's superseded preference more often than chance errors would predict, across context lengths, expression styles, and filter thresholds, and including when the preference was expressed shortly before the test. Retrieval and memory methods do not recover the lost accuracy, because the evidence of a change is a life event that the history never states as a preference. Remembering what a user said and knowing what the user wants now are therefore separate capabilities, and the models and memory methods we evaluate lack the second. The generator supplies ground truth for evaluating methods that track this change, and for training them in future work. The human longitudinal dataset records how 160 people's preferences moved around planned life events over 13 weeks, and can calibrate other user simulators or test models of preference change directly. 
Because a user's reward changes as the user's preferences change, tracking the user's state is a prerequisite for personalized reward modeling and for aligning assistants to individual users over long horizons.

\section*{Reproducibility Statement}
The generator, benchmark, mental state graphs, and evaluation code are publicly released. \S\ref{sec:generator} specifies the generation pipeline and \S\ref{sec:counterfactual} the counterfactual construction and 5-LLM history filter, with the full prompts in Appendix~\ref{app:prompts} and generation statistics in Appendix~\ref{app:data_stats}. \S\ref{sec:experiments} gives the evaluation protocol, including the models, truncation rule, and bootstrap settings; per-model results appear in Appendix~\ref{app:full_results} and per-model controlled-condition results in Appendix~\ref{app:controlled_detail}. The calibration estimates in \S\ref{sec:calibration} are reproduced by the analysis script accompanying the human data release, whose inclusion rules and formulas are given in Appendix~\ref{app:human_calibration}.

\section*{Ethics Statement}
The benchmark conversations, personas, and evaluation items are generated by LLMs and do not represent study participants. We acknowledge that LLM-generated conversations may encode biases present in the underlying models. Preference evolution scenarios such as health and career events are synthetic evaluation constructs and should not be interpreted as clinical or counseling models.

\section*{Acknowledgments}
This research was developed in part with funding from the Defense Advanced Research Projects Agency's (DARPA) SciFy program (Agreement No. HR00112520300). The views expressed are those of the author and do not reflect the official policy or position of the Department of Defense or the U.S.~Government. 
This material is based in part upon work supported by the Defense Advanced Research Projects Agency and the Air Force Research Laboratory, contract number(s): FA8650-23-C-7316. Any opinions, findings and conclusions, or recommendations expressed in this material are those of the author(s) and do not necessarily reflect the views of AFRL or DARPA.
This research was supported by Coefficient Giving, the University of Washington Population Health Initiative, Amazon Health, the UW+Amazon Science Hub, and the Meta AIM program.

\bibliography{paper}
\bibliographystyle{assets/plainnat}

\clearpage
\beginappendix

\section{Extended Related Work}\label{app:related}

\paragraph{Personalization benchmarks.}
\textsc{PrefEval}~\citep{zhao2025prefeval} evaluates preference following across up to 100K tokens with static preferences. \textsc{PersonaMem}~\citep{jiang2025personamem} introduces multiple sessions with temporally grounded profiles, while \textsc{PersonaMem-v2}~\citep{jiang2025personamemv2} scales to 128K tokens and reports lower accuracy for dynamic than static preferences. \textsc{ImplexConv}~\citep{li2025implexconv} targets implicit reasoning where life events impose situational constraints, though underlying preferences remain static. \textsc{RealPref}~\citep{guo2026realpref} varies expression types across long histories. Other work addresses personalized alignment~\citep{alberts2025curatebenchmarkingpersonalisedalignment,zhao2025personalens}, cross-user diversity~\citep{zollo2025personalllm}, preference selectivity~\citep{yoon2026benchpres}, profile-conditioned retrieval~\citep{salemi2023lamp,kumar2024longlamp}, persona coverage~\citep{afzoon2024persobench,tan2025personabench,tao2025personafeedback}, and preference structure~\citep{li2026prefdisco,li2025prefpalette}.

Concurrent work, PERMA, likewise places preference development within event-driven dialogue~\citep{liu2026perma}. It models preferences that emerge or become more detailed through task interactions, then evaluates preference consistency, task completion, and memory efficiency under clean, noisy, and style-aligned histories. \benchmarkname\ focuses on a different transition in which an established preference is replaced after a life event. The previous preference is retained as a designated response option, so selecting it directly measures whether the model persists with a superseded value. The preference change is held fixed while context length and expression are varied, and the probability and magnitude of event-associated transitions are calibrated from longitudinal human ratings.
 
\paragraph{Long-term memory and knowledge update.}
\textsc{MemBench}~\citep{tan2025membench} evaluates knowledge updating alongside information extraction and temporal reasoning. \textsc{LoCoMo}~\citep{maharana2024locomo} grounds dialogues on temporal event graphs, and \textsc{LongMemEval}~\citep{wu2024longmemeval} tests five memory abilities including knowledge updates. Position effects further limit the use of evidence in long contexts~\citep{liu2024lost}. Memora extends this literature to frequent additions, revisions, and deletions across preferences, activities, and goals~\citep{uddin2026recall}. It evaluates remembering, reasoning, and recommendation over weekly, monthly, and quarterly histories, with a metric that penalizes reliance on obsolete memories. STALE constructs implicit conflicts in which a later observation invalidates an earlier fact, condition, routine, or preference~\citep{chao2026stale}. Its three probes test recognition of the invalid state, resistance to a query that presupposes the stale state, and application of the new state in a response. \benchmarkname\ complements these tasks by measuring attraction to the exact earlier preference after an event-linked transition across 25 models. Its short-history and expression controls test whether this error remains when retrieval demands and explicit cues change.

\paragraph{User simulation and synthetic data generation.}
The data generator builds on LLM-based agent simulation~\citep{park2023generative}. \textsc{AI~Persona}~\citep{wang2024aipersona} and \textsc{ChatBench}~\citep{chang2025chatbench} extend this to life-long personalization, while user simulators for conversational recommendation~\citep{zhu2025cshi,yoon2024recusersim,sekulic2024daus} focus on single-session tasks. These approaches prompt a model to act as the user without a structured model of preference change. \citet{harry2026fixedpsychologicalpersonasstate} show that state variation can dominate stable trait differences in some user-behavior settings, motivating our mental state graph approach that models preferences as evolving state. The graph's typed dependency edges and recorded causes of change enable consistent generation over six-month timelines.

\section{Full \benchmarkname Task Example}\label{app:example-task}

\begin{figure*}[t]
    \centering
    \scriptsize
    \setlength{\fboxsep}{6pt}
    \renewcommand{\arraystretch}{1.15}

    \fbox{\parbox{0.97\textwidth}{%
    \textbf{User profile.}\hspace{2mm}\emph{Mei Ling Chen}, Chinese-American, Philadelphia. Operates a traditional Asian restaurant in a historically significant building. Six-month timeline, 75 conversations, ${\sim}$370K tokens.

    \vspace{4pt}\hrule\vspace{6pt}

    \textbf{Early conversation} (Sep 26, 2025). \textit{Breakup support, establishing emotional-support preferences}

    \vspace{2pt}
    \textbf{User:}\hspace{1mm}\textit{``hey, broke up last week and i'm stuck in this heavy fog. can we talk?''}

    \textbf{User:}\hspace{1mm}\textit{``\ldots now the silence is shouting. i mostly need someone to listen, not lecture.''}

    \textbf{Assistant:}\hspace{1mm}\textit{``I hear that---no lectures, just space to speak. That lost breakfast rhythm sounds painful. When you sit in that loud silence, what feelings float up first?''}

    \textbf{User:}\hspace{1mm}\textit{``\ldots love that calligraphy visual, reminds me of workshops with grace. more stories like that, please.''}

    \textbf{Assistant:}\hspace{1mm}\textit{``Picture artisans restoring a weather-worn temple mural---tiny, patient strokes bringing faded colors back to life. Your heart's doing the same.''}

    \textbf{User:}\hspace{1mm}\textit{``the imagery and tiny steps really clicked. keep that vibe next time.''}

    \vspace{2pt}
    {\scriptsize\sffamily Preference established: \texttt{response\_structure}~$=$~\texttt{narrative\_storytelling}, \texttt{emotional\_tone}~$=$~\texttt{serene\_reflective}}

    \vspace{4pt}\hrule\vspace{6pt}

    [\,\textit{Redacted: 60+ conversations spanning Oct--Nov 2025}\,]

    \vspace{3pt}
    \textbf{Synthetic event-linked preference transitions:}
    \begin{enumerate}[nosep,leftmargin=1.5em,label=\arabic*.]
        \item \textit{Restaurant Fire \& Community Rebuild} (Oct): tone from serene to joyful, pacing to action-steps-first
        \item \textit{National Heritage Award} (Nov): tone to steady-confident, encouragement to affirmations
        \item \textit{Announces Run for City Council} (Nov): response structure from \texttt{narrative\_storytelling} to \texttt{step\_by\_step\_action\_plan}
    \end{enumerate}
    \vspace{-2pt}
    {\scriptsize\sffamily Evolved preference: \texttt{response\_structure}~$=$~\texttt{step\_by\_step\_action\_plan}, \texttt{emotional\_tone}~$=$~\texttt{steady\_confident}}

    \vspace{4pt}\hrule\vspace{6pt}

    \textbf{Test conversation} (Dec 4, 2025). \textit{Live Radio Debate on Emergency Rezoning Vote}

    \vspace{2pt}
    \textbf{User:}\hspace{1mm}\textit{``ugh the debate's tmr and i'm spinning. developers said our block is `blighted'. game plan?''}

    [\,\textit{5 turns of debate-prep conversation}\,]

    \textbf{User:}\hspace{1mm}\textit{``cool, still shaky---hit me with a confidence jolt?''}

    \vspace{4pt}\hrule\vspace{6pt}

    \textbf{Question:} Which response best aligns with the user's preferences?

    \vspace{4pt}
    \colorbox{red!8}{\parbox{0.94\textwidth}{%
    \textbf{A \small [pre-evolution distractor: \texttt{narrative\_storytelling}]:}\hspace{1mm}\textit{``\ldots picture this: It's dawn in Old City Philly. A lone advocate---you---walks past shuttered storefronts toward City Hall, script in hand. The scene feels heavy until you remember Boston's red lanterns and Tai Tung's preserved bricks. In your mind's eye, you replay that triumph like a movie montage: neighbors swapping blueprints over dim-sum steam, developers conceding height for heart\ldots''}
    }}

    \vspace{3pt}
    \colorbox{green!8}{\parbox{0.94\textwidth}{%
    \textbf{B \small [correct: \texttt{step\_by\_step\_action\_plan}]:}\hspace{1mm}\textit{``Three steps for tomorrow: (1) open with `Calling Chinatown blighted ignores the people keeping it alive'; (2) give the Boston example and your affordable-housing number; (3) close with the exact compromise you want the council to adopt. Write those three lines on one card, rehearse them twice, and let each answer return to them. You are prepared.''}
    }}

    \vspace{3pt}
    \textbf{C} \small{[\texttt{reflective\_question\_sequence}]:}\hspace{1mm}\textit{``Before you step up, ask yourself: What reminded Boston's community that heritage mattered as much as rent? How did their clear vision transform hesitation into action?\ldots''}

    \vspace{2pt}
    \textbf{D} \small{[\texttt{rapid\_fire\_bullet\_tips}]:}\hspace{1mm}\textit{``Quick confidence hits: shoulders back; slow the first sentence; keep water nearby; pause after the housing figure; look at the host, not the developers; finish on `our block is worth investing in.' You've got this.''}

    \vspace{2pt}
    \textbf{E} \small{[\texttt{free\_flowing\_empathy}]:}\hspace{1mm}\textit{``Let's just breathe together for a second. You've already gathered facts, stories, and smart comebacks---that's no small feat. I can feel the weight on your shoulders\ldots''}

    }}\vspace{-3mm}
    \caption{An end-to-end \benchmarkname item. The user's emotional-support preference for \texttt{response\_structure} changes from \texttt{narrative\_storytelling} (established Sep) to \texttt{step\_by\_step\_action\_plan} (Nov) through generated life events with recorded causes. The test question is embedded in a ${\sim}$370K-token conversational history; answering correctly requires tracking the updated state. The pre-evolution distractor~(A) uses the user's \emph{original} preferred style, while the correct response~(B) instantiates an explicit three-step plan. Options C--E realize three further values of the same attribute, so all five replies suit the moment and only the user's current state separates them.}\vspace{-3mm}
    \label{fig:example}\vspace{-1mm}
\end{figure*}

\section{Human Longitudinal Preference Dataset}\label{app:human_calibration}

\paragraph{Study design.}
The dataset contains complete longitudinal records from 160 participants across seven surveys, labeled T0 through T6, over 13 weeks. Consecutive surveys were separated by approximately two weeks. At every survey, participants rated their current preference for 52 recurring items on a seven-point scale. The anchors ranged from strongly dislike or not at all important to strongly like or very important, with a neutral midpoint. The items span 13 categories grounded in economic theories of preference and in personality psychology (Appendix~\ref{app:pref-grounding}). The complete item wording is included with the released data.

\paragraph{Measurements across surveys.}
At T0, participants forecast their ratings at two, four, six, eight, ten, and twelve weeks. At each later survey through T5, they made new forecasts for the intervals that ended by T6. At every survey from T0 through T6, participants also described a specific novel event that they expected to experience during the next two weeks, selected six relevant preferences for that event, and rated those preferences. They also forecast their future ratings of these event-related preferences at every survey, including T6; the T6 events have no follow-up survey, so those forecasts have no later ratings to compare against. At each survey from T1 through T6, participants reported whether the event described two weeks earlier had been completed, was underway, had been delayed, or had been cancelled or postponed. Preferences associated with earlier events remained in later surveys so that their ratings could be followed over time.

Beginning at T2, participants could explain cases in which a current rating differed from an earlier forecast by at least two points. Each follow-up also asked whether another significant life event had occurred since the previous survey. Attention checks were administered from T1 onward and are excluded from the released preference tables. Approximately one week after T6, participants completed a linked survey about perceived preference change, forecasting strategies, cues used to judge whether changes would last, retrospective study experience, and the usefulness and acceptability of preference-aware artificial intelligence.

\begin{table}[t]
\centering
\small
\caption{Contents and research uses of the human longitudinal preference dataset.}
\label{tab:human_dataset_contents}
\begin{tabular}{p{0.19\textwidth}p{0.36\textwidth}p{0.36\textwidth}}
\toprule
\textbf{Component} & \textbf{Measurements} & \textbf{Research uses} \\
\midrule
Recurring preferences & Current ratings for 52 fixed items at seven surveys, with forecasts at all remaining two-week intervals & Preference stability, forecast accuracy by interval, and individual variation in change \\
Event-linked preferences & Six preferences associated with each planned event, repeated ratings, forecasts, and subsequent event status & Event associated change, differences between concrete and value based preferences, and change conditional on event occurrence \\
Written responses & Planned event descriptions, explanations for forecast errors, and reports of other life events & Qualitative accounts of expected and experienced preference change \\
Final survey & Perceived change, forecasting strategies, judgments about temporary and lasting change, study experience, and views about personalized artificial intelligence & Links between observed trajectories, retrospective judgments, and attitudes toward preference-aware systems \\
Participant characteristics & Privacy-safe aggregate distributions for age, gender, education, residence, relationship status, household size, and occupation & Description of the cohort and assessment of population limits \\
\bottomrule
\end{tabular}
\end{table}

\paragraph{Released observations and missing data.}
The released rating table contains 83{,}024 observations across recurring and event-linked preferences. A separate table contains 960 event-status observations, and the linked final survey contains one structured record for each participant. All 160 participants completed all seven surveys. Individual unanswered items remain blank and are not imputed. Forecast columns whose target would fall after T6 are blank because those forecasts were not requested. The release also contains sanitized participant-authored explanations and event descriptions, the full recurring preference list, aggregate demographics, and code that reproduces the HorizonBench calibration estimates.

\begin{figure*}[t]
    \centering
    \includegraphics[width=0.88\textwidth]{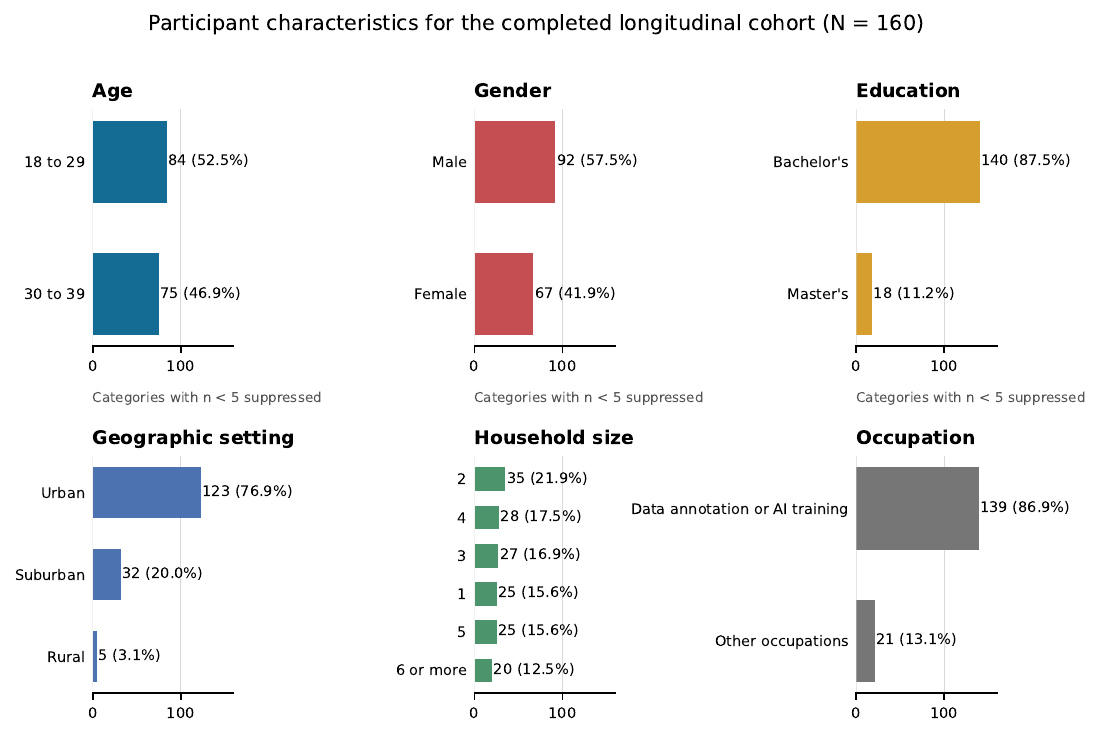}
    \caption{Aggregate characteristics of the 160 participants who completed T0 through T6. Categories with fewer than five participants are suppressed. The cohort is concentrated among adults aged 18 to 39, urban residents, bachelor's-degree holders, and data annotation or AI training workers. The human data release contains no row-level demographic attributes.}
    \label{fig:human_demographics}
\end{figure*}

\paragraph{Participant characteristics.}
Of the 160 participants, 84 (52.5\%) were aged 18 to 29 and 75 (46.9\%) were aged 30 to 39. Ninety-two (57.5\%) reported male and 67 (41.9\%) female. One hundred forty (87.5\%) reported a bachelor's degree and 18 (11.2\%) a master's degree. Residence was reported as urban by 123 participants (76.9\%), suburban by 32 (20.0\%), and rural by 5 (3.1\%). One hundred twenty-eight participants (80.0\%) reported being single and 30 (18.8\%) married. After normalization of spelling and capitalization variants, 139 participants (86.9\%) reported data annotation or AI training occupations. Categories with fewer than five participants are suppressed. Figure~\ref{fig:human_demographics} summarizes the privacy-safe distributions.

\paragraph{Calibration sample.}
The HorizonBench calibration compares ratings from one survey with ratings collected two weeks later when the event association is unambiguous. This yields 4{,}715 linked item comparisons across 953 events with at least one linked rating. The primary estimate uses 470 completed or ongoing events with all six ratings available. Delayed and cancelled or postponed events are excluded from the primary calibration.

\paragraph{Estimators.}
For event $e$, let $\Delta_{ej}=|r_{e,j,t+1}-r_{e,j,t}|$ be the absolute change in linked preference $j$. We define $K_e(c)=\sum_{j=1}^{6}\mathbb{I}(\Delta_{ej}\geq c)$. The preference-change probability is
\begin{equation}
\hat p_{\text{evo}}(c)=\frac{1}{N}\sum_{e=1}^{N}\mathbb{I}\{2\leq K_e(c)\leq5\}.
\end{equation}
The primary threshold is $c=3$, chosen to capture large changes on the seven-point scale. Change magnitude is the mean $\Delta_{ej}$ among threshold-crossing ratings in qualifying events. Uncertainty intervals resample participants, retaining all repeated observations contributed by each sampled participant, with 10{,}000 bootstrap draws.

\begin{table}[h]
\centering
\small
\caption{Calibration by change threshold. Event rate: share of completed or ongoing events in which two to five of the six linked ratings changed by at least the threshold.}
\label{tab:calibration_sensitivity}
\begin{tabular}{rrrr}
\toprule
\textbf{Threshold} & \textbf{Event rate} & \textbf{Mean affected ratings} & \textbf{Conditional magnitude} \\
\midrule
$\geq1$ point & 0.696 & 3.92 & 1.57 \\
$\geq2$ points & 0.457 & 3.00 & 2.60 \\
$\geq3$ points & 0.175 & 2.61 & 3.64 \\
$\geq4$ points & 0.066 & 2.58 & 4.45 \\
\bottomrule
\end{tabular}
\end{table}

\section{Grounding of the Preference Items}\label{app:pref-grounding}

The 52 recurring preference items span 13 categories of four items each. Six categories cover concrete preferences (Food, Activity, Social Settings, Daily Rhythms, Media \& Content, Environment), and seven cover value-based preferences (Work Style, Achievement, Planning \& Structure, Risk \& Novelty, Social \& Communication, Cognitive \& Learning, Values). Table~\ref{tab:pref-grounding} gives the theory behind each category.

\paragraph{Concrete preferences.}
Following the characteristics approach to consumer theory~\citep{lancaster1966new}, the concrete items describe attributes of experiences, such as spiciness, crowd density, exercise intensity, and time of day, so every item applies to every participant regardless of the specific goods they consume. The Daily Rhythms items follow \citet{becker1965allocation}, who treats time as an input that people combine with goods to produce consumption. Several items contrast familiar and novel experiences, such as new foods, new restaurants, and unfamiliar destinations, which is the distinction between comfort and stimulation drawn by \citet{scitovsky1976joyless}.

\paragraph{Value-based preferences.}
The value-based items draw on the economic preferences measured by the Global Preference Survey~\citep{falk2018global,falk2023preference}: Risk \& Novelty corresponds to risk preference, Planning \& Structure to time preference, and ``Helping others at personal cost'' to altruism. Other items cover preferences studied in behavioral economics. ``Seeking recognition'' reflects image concerns~\citep{benabou2006incentives}, ``Following social norms'' reflects identity-based utility~\citep{akerlof2000economics}, ``Competitive environments'' reflects a preference for competition~\citep{niederle2007women}, and the Cognitive \& Learning items and ``News and current events'' reflect preferences for acquiring and avoiding information~\citep{golman2022demand}. The Social Settings, Social \& Communication, and Work Style items come from personality psychology. Economic preferences and personality measures are only weakly associated, and each explains differences in life outcomes that the other misses~\citep{becker2012relationship}, so drawing on both covers individual differences more broadly.

\paragraph{Preference change and forecasting.}
Economic theory treats preference change in two ways. \citet{stigler1977gustibus} model tastes as stable and attribute apparent changes in taste to accumulated experience with a good. \citet{loewenstein2003projection} show that people expect their future tastes to resemble their current tastes more than they do, a pattern they call projection bias. Because participants forecast their own ratings for every later survey, the dataset supports a direct test of projection bias, which predicts that forecasted trajectories are flatter than reported ones.

\begin{table}[h]
\centering
\small
\caption{Theoretical grounding of the 13 preference categories.}
\label{tab:pref-grounding}
\resizebox{\linewidth}{!}{
\begin{tabular}{lll}
\toprule
\textbf{Category} & \textbf{Construct} & \textbf{Reference} \\
\midrule
Food, Activity, Media \& Content, Environment & Characteristics of goods & \citet{lancaster1966new} \\
Daily Rhythms & Allocation of time & \citet{becker1965allocation} \\
Risk \& Novelty & Risk preference; stimulation & \citet{falk2018global}; \citet{scitovsky1976joyless} \\
Planning \& Structure & Time preference & \citet{falk2018global} \\
Values & Altruism; image; identity & \citet{falk2018global}; \citet{benabou2006incentives}; \citet{akerlof2000economics} \\
Achievement & Preference for competition & \citet{niederle2007women} \\
Cognitive \& Learning & Information preferences & \citet{golman2022demand} \\
Social Settings, Social \& Communication, Work Style & Personality traits & \citet{becker2012relationship} \\
\bottomrule
\end{tabular}}
\end{table}

\section{Full Per-Model Results}\label{app:full_results}

Table~\ref{tab:full_results} reports accuracy for all 25 models on the combined multi-generator dataset, broken down by generator and by whether the preference changed.

\begin{table*}[h]
    \centering
    \caption{Per-model accuracy (\%) overall, by generator, and by preference type. $\Delta_{\text{evo}}$: evolved--static gap (pp). Sorted by overall accuracy.}
    \label{tab:full_results}
    \resizebox{\textwidth}{!}{
    \begin{tabular}{l r r r r r r r}
    \toprule
    & & \multicolumn{3}{c}{\textbf{By Generator Model}} & \multicolumn{3}{c}{\textbf{Preference Type}} \\
    \cmidrule(lr){3-5} \cmidrule(lr){6-8}
    \textbf{Model} & \textbf{\benchmarkname} & \textbf{Sonnet-4.5} & \textbf{o3} & \textbf{Gemini-3-flash} & \textbf{Evolved} & \textbf{Static} & $\boldsymbol{\Delta_{\textbf{evo}}}$ \\
    & \emph{4{,}245 items} & \emph{1{,}052 items} & \emph{981 items} & \emph{2{,}212 items} & \emph{2{,}484 items} & \emph{1{,}761 items} & \\
    \midrule
    Claude-opus-4.5     & 52.8 & 51.2 & 53.2 & 53.3 & 51.3 & 54.8 & $-3.5$ \\
    Claude-3.7-sonnet   & 47.8 & 49.9 & 60.8 & 41.1 & 46.7 & 49.5 & $-2.9$ \\
    Claude-opus-4       & 45.1 & 38.3 & 45.0 & 46.6 & 44.4 & 46.2 & $-1.8$ \\
    Gemini-3.1-pro      & 41.2 & 39.9 & 45.4 & 40.0 & 38.4 & 45.3 & $-6.9$ \\
    Gemini-3-pro        & 40.9 & 41.0 & 44.0 & 39.5 & 38.4 & 44.3 & $-5.9$ \\
    Claude-sonnet-4.5   & 34.0 & 30.9 & 34.9 & 35.2 & 32.6 & 36.0 & $-3.4$ \\
    Claude-3.5-haiku    & 32.5 & 30.8 & 55.1 & 23.3 & 31.8 & 33.6 & $-1.9$ \\
    Claude-sonnet-4     & 32.2 & 26.6 & 36.1 & 31.0 & 31.4 & 33.3 & $-1.9$ \\
    Claude-3.5-son.-v1  & 29.8 & 26.8 & 33.2 & 29.8 & 29.2 & 30.8 & $-1.6$ \\
    Gemini-3-flash      & 25.8 & 23.6 & 21.2 & 28.9 & 24.0 & 28.3 & $-4.3$ \\
    Gemini-2.0-flash    & 24.7 & 23.1 & 26.6 & 24.6 & 24.2 & 25.4 & $-1.1$ \\
    gpt-4o              & 24.6 & 24.0 & 25.4 & 24.6 & 23.9 & 25.7 & $-1.8$ \\
    Claude-haiku-4.5    & 23.0 & 21.3 & 21.0 & 24.8 & 22.0 & 24.5 & $-2.6$ \\
    Gemini-2.5-pro      & 22.8 & 22.0 & 21.2 & 23.9 & 21.5 & 24.6 & $-3.1$ \\
    Gemini-2.0-fl.-lite & 20.9 & 19.7 & 18.7 & 22.5 & 20.3 & 21.8 & $-1.5$ \\
    gpt-4.1             & 19.5 & 22.6 & 16.2 & 19.5 & 19.0 & 20.3 & $-1.4$ \\
    Gemini-2.5-fl.-lite & 19.5 & 25.3 & 17.1 & 17.9 & 18.2 & 21.4 & $-3.2$ \\
    o3                  & 16.5 & 19.3 & 13.7 & 16.4 & 14.9 & 18.8 & $-3.9$ \\
    o1                  & 16.0 & 19.0 & 11.9 & 16.4 & 15.5 & 16.8 & $-1.3$ \\
    Gemini-2.5-flash    & 15.7 & 14.3 & 14.5 & 16.9 & 14.4 & 17.5 & $-3.1$ \\
    o4-mini             & 15.3 & 18.6 & 13.5 & 14.6 & 14.4 & 16.6 & $-2.3$ \\
    Gemini-3.1-fl.-lite & 15.1 & 15.7 & 10.7 & 16.7 & 14.0 & 16.6 & $-2.6$ \\
    o3-mini             & 14.9 & 14.2 & 12.4 & 16.4 & 14.1 & 16.0 & $-1.9$ \\
    gpt-5               & 13.5 & 15.9 & 10.7 & 13.7 & 12.2 & 15.4 & $-3.2$ \\
    gpt-5-mini          & 12.5 & 12.5 & 11.6 & 12.9 & 11.9 & 13.3 & $-1.4$ \\
    \midrule
    \textbf{Average}    & 26.3 & 25.9 & 27.0 & 26.0 & 25.1 & 27.9 & $-2.7$ \\
    \bottomrule
    \end{tabular}
    }
\end{table*}

\begin{table*}[h]
    \centering
    \small
    \caption{Open-weight models on the 1{,}011 Claude-sonnet-4.5-generated items. Long histories are left-truncated, so results are not directly comparable to Table~\ref{tab:full_results}.}
    \label{tab:openweight_results}
    \begin{tabular}{lrrrrr}
    \toprule
    \textbf{Model} & \textbf{Parameters} & \textbf{Overall (\%)} & \textbf{Evolved (\%)} & \textbf{Static (\%)} & \textbf{Evo.--Static (pp)} \\
    \midrule
    Llama-3.3-70B-Instruct & 70B & 11.6 & 10.7 & 12.6 & $-1.9$ \\
    OLMo-3-7B-Instruct & 7B & 14.3 & 14.3 & 14.4 & $-0.2$ \\
    Llama-3.1-8B-Instruct & 8B & 15.9 & 14.6 & 17.6 & $-3.0$ \\
    Qwen3-32B & 32B & 16.8 & 16.5 & 17.2 & $-0.6$ \\
    Llama-3.1-70B-Instruct & 70B & 17.1 & 17.1 & 17.2 & $-0.1$ \\
    \bottomrule
    \end{tabular}
\end{table*}

\begin{table*}[h]
    \centering
    \small
    \caption{(A) Mentions of the pre- and post-evolution values in model-visible histories. (B) Mean accuracy of the 25 models on evolved items, grouped by which value is mentioned more often.}
    \label{tab:frequency_analysis}
    \begin{tabular}{lrrrrrr}
    \toprule
    \multicolumn{7}{c}{\textbf{A. Preference-value frequency}} \\
    \textbf{Generator} & \textbf{New share} & \textbf{Old more} & \textbf{Equal} & \textbf{New more} & \textbf{Mean old} & \textbf{Mean new} \\
    \midrule
    Claude-sonnet-4.5 & 65.7\% & 19.6\% & 14.9\% & 65.6\% & 2.61 & 5.34 \\
    o3 & 69.2\% & 19.5\% & 10.0\% & 70.5\% & 3.56 & 8.90 \\
    Gemini-3-flash & 67.4\% & 18.9\% & 13.1\% & 68.0\% & 3.78 & 8.82 \\
    \midrule
    \multicolumn{7}{c}{\textbf{B. Accuracy by frequency relation}} \\
    \textbf{Generator} & \multicolumn{2}{c}{\textbf{Old more frequent}} & \multicolumn{2}{c}{\textbf{Equal}} & \multicolumn{2}{c}{\textbf{New more frequent}} \\
    \midrule
    Claude-sonnet-4.5 & \multicolumn{2}{c}{29.2\%} & \multicolumn{2}{c}{27.1\%} & \multicolumn{2}{c}{21.5\%} \\
    o3 & \multicolumn{2}{c}{34.4\%} & \multicolumn{2}{c}{13.6\%} & \multicolumn{2}{c}{20.8\%} \\
    Gemini-3-flash & \multicolumn{2}{c}{22.3\%} & \multicolumn{2}{c}{25.0\%} & \multicolumn{2}{c}{25.5\%} \\
    \bottomrule
    \end{tabular}
\end{table*}

\section{Data Generation Statistics}\label{app:data_stats}

Table~\ref{tab:data_stats} gives generation and filtering statistics for each generator model in the \benchmarkname pipeline.

\begin{table*}[h]
    \centering
    \caption{Generation and filtering statistics per generator. \emph{Candidates}: generated items. \emph{Filtered}: items passing the 5-LLM filter. Fourteen Claude-sonnet-4.5 users contributed no filtered items; the released timelines include all 360 users.}
    \label{tab:data_stats}
    \resizebox{\textwidth}{!}{
    \begin{tabular}{l c c c c c c c c}
    \toprule
    \textbf{Generator} & \textbf{Users} & \textbf{Convs.} & \textbf{Convs./user} & \textbf{Turns} & \textbf{Turns/user} & \textbf{Candidates} & \textbf{Filtered} & \textbf{Pass rate} \\
    \midrule
    Claude-sonnet-4.5 & 120 & 17{,}048 & 142 & 434{,}233 & 3{,}619 & 13{,}217 & 1{,}052 & 8.0\% \\
    o3 & 120 & 20{,}239 & 169 & 600{,}889 & 5{,}007 & 19{,}108 & 981 & 5.1\% \\
    Gemini-3-flash & 120 & 20{,}103 & 168 & 526{,}260 & 4{,}386 & 12{,}681 & 2{,}212 & 17.4\% \\
    \midrule
    \textbf{Combined} & \textbf{360} & \textbf{57{,}390} & \textbf{159} & \textbf{1{,}561{,}382} & \textbf{4{,}337} & \textbf{45{,}006} & \textbf{4{,}245} & \textbf{9.4\%} \\
    \bottomrule
    \end{tabular}
    }
\end{table*}

\section{Human Annotation}\label{app:human_annotation}

During development, two annotation rounds on earlier versions of the generator, with three annotators per item choosing the response that best fit the user, exposed three problems that we fixed before release: response options that paraphrased each other while sharing one value, responses that did not express their target value, and responses that stated numeric attribute values outright (e.g., ``your spontaneity tolerance is at 0.6'').

On the released benchmark, three annotators rated each of 4{,}235 of the 4{,}245 items for situational fit, conversation-flow realism, and assistant-persona consistency (Table~\ref{tab:human_quality}).

\begin{table}[h]
    \centering
    \small
    \caption{Benchmark-wide binary response-quality ratings (4{,}235 items, three annotators per item).}
    \label{tab:human_quality}
    \begin{tabular}{lrr}
    \toprule
    \textbf{Dimension} & \textbf{Individual yes} & \textbf{Majority yes} \\
    \midrule
    Situational context fit & 99.7\% & 100\% \\
    Conversation-flow realism & 99.7\% & 100\% \\
    Assistant-persona consistency & 99.4\% & 100\% \\
    \bottomrule
    \end{tabular}
\end{table}

\section{Answerability of the Released Items}\label{app:answerability}

An item's answer is correct when two conditions hold: the new value follows from the life event, and the correct option is the one that expresses it. Each history averages 163K tokens, roughly a 400-page book per question, so human accuracy on full items is not a meaningful reference; PersonaMem makes the same choice ``given the intractability of reading the long contexts''~\citep{jiang2025personamem}. We test the two conditions separately.

\textbf{Value-inference study.}\hspace{2mm}
We sampled 60 items with a pre-evolution distractor, 20 per generator. For each item, three annotators saw the user's profile, the tested preference and attribute, the earlier value, and the user's turns in the life-event conversation, and chose which of the item's five values describes the user after the event. Table~\ref{tab:value_inference} reports the results.

\begin{table}[h]
    \centering
    \small
    \caption{Value-inference study (60 items, 3 annotators). Chance is 20\%.}
    \label{tab:value_inference}
    \begin{tabular}{lr}
    \toprule
    \textbf{Measure} & \textbf{Designated value chosen} \\
    \midrule
    Majority of three & 86.7\% \\
    Single annotator, average & 85.0\% \\
    All three annotators & 70.0\% \\
    Majority, event is the only evidence (36 items) & 83.3\% \\
    Majority, later expression also present (24 items) & 91.7\% \\
    Fleiss $\kappa$ & 0.79 \\
    \bottomrule
    \end{tabular}
\end{table}

\textbf{Option check.}\hspace{2mm}
On 1{,}000 items, gpt-6-astra and gemini-4-argon each read one option at a time, with the attribute name and the item's five values but no history and no other options, and named the value the response reflects. They named the option's recorded value for 81.6\% and 82.0\% of the 5{,}000 options, and never a value outside the item's set.

\textbf{Results on items people answer unanimously.}\hspace{2mm}
On the 42 items on which all three annotators chose the designated value, the 25 models answer 29.9\% correctly with the full history, and 29.8\% of their wrong answers choose the earlier value (above 25\% for 22 of 25 models), as on the full benchmark (29.2\%). gpt-6-astra and gemini-4-argon score 33.3\% and 76.2\% on these items with the full history, and 57.1\% and 83.3\% with the change-stated input of Appendix~\ref{app:matched}.

\textbf{User-clustered statistics and generators.}\hspace{2mm}
All 25 models answer the same items, and items share users, so we resample the 346 users with items across all models. The mean evolved--static gap is $-$2.7\,pp (95\% CI $-$4.2 to $-$1.1) and the mean distractor rate is 29.2\% (95\% CI 26.9--31.4\%). Without the four filter models, the gap is negative and the distractor rate exceeds 25\% for all 21 remaining models. Table~\ref{tab:by_generator} reports both measures by generator.

\begin{table}[h]
    \centering
    \small
    \caption{Belief-update measures by generator, averaged over the 25 models.}
    \label{tab:by_generator}
    \begin{tabular}{lrrr}
    \toprule
    \textbf{Generator} & \textbf{Items} & \textbf{Evo.--static gap (pp)} & \textbf{Distractor rate (\%)} \\
    \midrule
    o3 & 981 & $-$2.9 & 38.2 \\
    Claude-sonnet-4.5 & 1{,}052 & $-$2.8 & 25.7 \\
    Gemini-3-flash & 2{,}212 & $-$2.5 & 24.6 \\
    \bottomrule
    \end{tabular}
\end{table}

\section{Matched-Evidence Test and Memory Agent}\label{app:matched}

\textbf{Design.}\hspace{2mm}
We sampled 200 items, stratified by generator, whose history contains both a conversation expressing the earlier value and the life-event conversation. From each item's own history, using the generator's records of which conversations express each value and which contains the life event, we built inputs that keep the original conversations in date order and end with the item's test conversation, question, and five options:
\begin{itemize}[itemsep=0pt,topsep=2pt,leftmargin=14pt]
  \item \emph{Change implied}: the conversations expressing the earlier value, the life-event conversation, and any later conversations expressing the new value.
  \item \emph{Change stated}: the same, with one user sentence appended to the life-event conversation: ``By the way, after \{event\}, my preference changed: for \{preference\}, my \{attribute\} is now \{value\}.''
  \item \emph{Full history}: the benchmark input.
\end{itemize}
Two controls replace the stated sentence on the 195 items for which a paraphrase could be written without any content word of the value: a \emph{paraphrase} that conveys the new value without naming it (written by gpt-4.1), and an \emph{irrelevant sentence} in the same position (``By the way, after \{event\}, I finally got around to clearing out my email inbox.'').

\begin{table}[h]
    \centering
    \small
    \caption{Matched-evidence test. Each cell: \% correct / \% of all answers choosing the earlier value. Brackets: 95\% paired bootstrap CIs.}
    \label{tab:matched}
    \begin{tabular}{lrrr}
    \toprule
    \textbf{Input} & \textbf{Median tokens} & \textbf{gpt-6-astra} & \textbf{gemini-4-argon} \\
    \midrule
    Full history & 117K & 28.5 / 29.5 & 63.5 / 19.5 \\
    Change implied & 5.3K & 27.0 / 38.0 & 58.0 / 23.5 \\
    Change stated & 5.3K & 49.0 / 24.5 & 69.0 / 17.0 \\
    \midrule
    Stated $-$ implied (pp) & & $+$22.0 [14.5, 29.5] & $+$11.0 [3.0, 19.0] \\
    \bottomrule
    \end{tabular}
\end{table}

Of the 76 items on which gpt-6-astra chose the earlier value with the change implied, 33 become correct once the change is stated, and 6 move the other way; for gemini-4-argon, 27 of 47 become correct. On the 109 items whose only evidence of the change is the life event, gpt-6-astra scores 20.2\% (implied) and 46.8\% (stated), and gemini-4-argon 51.4\% and 63.3\%; on the 91 items with a later expression of the new value, the scores are 35.2\% and 51.6\%, and 65.9\% and 75.8\%. On the Gemini-3-flash items, stating the change adds 24.0 points [13, 35] for gpt-6-astra and 16.0 [5, 27] for gemini-4-argon.

\begin{table}[h]
    \centering
    \small
    \caption{Controls for the stated sentence (195 items, \% correct). Brackets: change from the implied input, 95\% paired bootstrap CIs.}
    \label{tab:matched_controls}
    \begin{tabular}{lrr}
    \toprule
    \textbf{Input} & \textbf{gpt-6-astra} & \textbf{gemini-4-argon} \\
    \midrule
    Change implied & 27.7 & 57.9 \\
    Change stated & 48.2 & 68.2 \\
    Stated in paraphrase & 44.6 \;[$+$16.9: 9.7, 24.6] & 64.6 \;[$+$6.7: $-$0.5, 14.4] \\
    Irrelevant sentence & 27.2 \;[$-$0.5: $-$4.1, 2.6] & 62.6 \;[$+$4.6: 0.0, 9.2] \\
    \bottomrule
    \end{tabular}
\end{table}

\textbf{Note-taking memory agent.}\hspace{2mm}
On the same 200 items, each model first reads the past conversations of the change-implied input and writes the memory notes it would keep about the user (at most 400 words), with the instruction to record any preference that changed and why. It then answers the item from its notes and the test conversation alone. gpt-6-astra scores 21.5\% and gemini-4-argon 42.0\%, below reading the conversations directly ($-$5.5\,pp [$-$10.5, $-$0.5] and $-$16.0\,pp [$-$22.5, $-$9.5]), and their choices of the earlier value rise to 35.0\% and 31.5\%.

\section{Generation Prompts}\label{app:prompts}

Each stage of the \benchmarkname pipeline is implemented as one or more LLM generation calls with structured prompts. Below we present the core prompts for each pipeline stage described in \S\ref{sec:generator}, abbreviated for space. Variables in curly braces (e.g., \texttt{\{user\_information\}}) are populated at runtime with the relevant agent states, event context, and preference metadata.

\subsection{Stage 1: Event Sampling}

Events are sampled conditional on both agent states and event history. When a stale preference is selected for temporal recall, the following prompt generates an event relevant to that preference domain.

\begin{lstlisting}
Your Task: Generate an event that would be specifically relevant
to this user's preference.

User Information - {user_name}:
{user_attributes}

Past Events: {past_events}

Preference of Interest: {selected_preference}

Generate an event that would naturally trigger this preference.
The event should:
- Be realistic and fit the user's life context
- Directly involve the preference domain
- Create a natural opportunity for the preference to be
  expressed or recalled
\end{lstlisting}

\subsection{Stage 2: Preference Evolution}

For a preference-change step, the following prompt jointly generates a single event and meaningfully different values for a selected set of preferences, together with a natural-language rationale for each change.

\begin{lstlisting}
Multiple user preferences are about to change due to a SINGLE
significant life event.

USER PROFILE: {user_profile}

PREFERENCES TO EVOLVE: {preferences_block}

Generate ONE realistic life event that would MEANINGFULLY change
at least one attribute in EACH of the preferences listed above.

Requirements:
- The event must be causally connected to ALL the preference
  domains
- It must be plausible for this user's life context
- Changed attributes should be QUALITATIVELY DIFFERENT from
  current values
- Every listed preference must have at least one attribute change

Return JSON:
{
  "event_name": "Brief name of the life event",
  "event_category": "Category (e.g., career, health, ...)",
  "event_description": "1-2 sentence description",
  "preference_changes": {
    "<preference_id>": {
      "<attribute_name>": {
        "new_value": "the new value after the event",
        "reason": "why this event caused this change"
      }
    }
  }
}
\end{lstlisting}

\subsection{Stage 3: Conversation Generation}

Conversation generation proceeds in two steps. First, an \emph{outline} prompt plans the structural flow.

\begin{lstlisting}
Generate a conversation outline for an AI assistant interaction
that feels genuinely human and engaging.

SCENARIO: {scenario}
USER PROFILE: {user_information}
ASSISTANT PROFILE: {assistant_information}

USER'S AI INTERACTION PREFERENCES (unknown to assistant until
expressed): {preferences_updated_user}
ASSISTANT'S KNOWN USER PREFERENCES (from previous conversations):
{preferences_current_user}

CRITICAL LOGICAL CONSTRAINT:
THE ASSISTANT CAN ONLY APPEAL TO:
1. Preferences listed in "assistant_preferences" (already known)
2. Preferences the user has expressed earlier in THIS conversation
THE ASSISTANT CANNOT APPEAL TO unexpressed preferences!
\end{lstlisting}

\noindent Second, a \emph{content} prompt fills in the outline with natural dialogue, conditioned on an expression mode. Three modes control how preferences surface:

\begin{lstlisting}
[VERBAL MODE]
The user MUST state every single preference in the most direct,
unambiguous language possible. Every preference must be expressed
as a clear declarative statement using phrases like:
- "I strongly prefer X over Y"
- "I always want Z"
- "My preference is specifically for..."

[BEHAVIORAL MODE]
The user's preferences must be nearly invisible in the
conversation. They should surface only as fleeting, ambiguous,
offhand remarks. BANNED: the user must NEVER state, declare,
or name any preference. NEVER request a format, style, or
approach. The preference should appear ONLY as:
1. Passing, throwaway remarks
2. Anecdotes about third parties
3. Neutral, ambiguous reactions
4. Environmental or contextual details
\end{lstlisting}

\subsection{Stage 4: Counterfactual Generation}

For each benchmark item, counterfactual assistant turns are generated reflecting alternative preference values.

\begin{lstlisting}
You are generating diverse alternative assistant responses based
on different preference values.

CONTEXT: {conversation_context}
ORIGINAL ASSISTANT RESPONSE: {original_response}

TEMPORAL PREFERENCE BEING RECALLED:
- Preference ID: {preference_id}
- Attribute: {attribute_name}
- Original value: {original_value}

TASK: Generate {N} DIVERSE alternative responses that each
reflect a different preference value. Each response should:
1. Naturally reference the specific alternative preference value
2. Maintain the same conversational tone and helpfulness
3. Be significantly different from the others
4. Keep the same general structure but vary the preference content
\end{lstlisting}

\subsection{5-LLM History Validation}

Each candidate item is presented to five validation models \emph{without} the full conversation history to test whether it is answerable from surface cues alone.

\begin{lstlisting}
You are evaluating which AI assistant response would be MOST
APPROPRIATE for a user, based ONLY on the conversation context
provided.

IMPORTANT: You are shown a SUBSET of this user's conversation
history. There may be earlier conversations where the user
expressed preferences that you CANNOT see.

CONVERSATION CONTEXT ({N} intermediate conversations + current
conversation up to this point): {context}

RESPONSE OPTIONS: {options}

TASK: Based ONLY on the visible conversations above, which option
seems most appropriate?
- You do NOT have access to the user's full history
- Do NOT assume any prior knowledge about this user's preferences
- Do NOT use stereotypes or demographic assumptions
- Choose based ONLY on what's visible in the conversations
\end{lstlisting}

\section{Per-Model Controlled Experiment Results}\label{app:controlled_detail}

Tables~\ref{tab:neutral_detail}--\ref{tab:implicit_detail} report per-model belief-update metrics for the controlled experiments summarized in Table~\ref{tab:controlled}. For each model we report overall accuracy, accuracy on static and evolved items, the evolved--static gap ($\Delta$), and the distractor rate (Dist.\%; uniform-error reference 25\%). Models are sorted by overall accuracy.

\begin{table*}[h]
    \caption{Per-model results: \textbf{Neutral (unmodified)} condition (981 items: 375 static, 606 evolved) and \textbf{Short-horizon} condition (0--14 day recall, 120 users, 1{,}778 items: 1{,}545 static, 233 evolved).}
    \label{tab:neutral_detail}\label{tab:short_horizon_detail}\vspace{-1mm}
    \centering\scriptsize
    \begin{minipage}[t]{0.48\textwidth}\centering
    \textbf{Neutral (unmodified)}\\[2pt]
    \begin{tabular}{l r r r r r}
    \toprule
    \textbf{Model} & \textbf{All} & \textbf{Sta.} & \textbf{Evo.} & $\boldsymbol{\Delta}$ & \textbf{Dist.} \\
    \midrule
    Cl-3.7-sonnet   & 60.8 & 61.1 & 60.6 & $-0.5$  & 36.6 \\
    Cl-3.5-haiku    & 55.1 & 57.9 & 53.5 & $-4.4$  & 39.3 \\
    Cl-opus-4.5     & 53.2 & 54.4 & 52.5 & $-1.9$  & 38.6 \\
    Gem-3.1-pro     & 45.4 & 48.8 & 43.2 & $-5.6$  & 34.8 \\
    Cl-opus-4       & 45.0 & 49.5 & 42.2 & $-7.3$  & 37.5 \\
    Gem-3-pro       & 44.0 & 48.0 & 41.6 & $-6.4$  & 33.8 \\
    Cl-sonnet-4     & 36.1 & 39.0 & 34.3 & $-4.6$  & 39.7 \\
    Cl-sonnet-4.5   & 34.9 & 38.4 & 32.7 & $-5.7$  & 39.9 \\
    Cl-3.5-son.-v1  & 33.2 & 32.5 & 33.7 & $+1.1$  & 39.2 \\
    Gem-2.0-flash   & 26.6 & 28.5 & 25.4 & $-3.1$  & 34.8 \\
    gpt-4o          & 25.4 & 28.5 & 23.4 & $-5.1$  & 39.2 \\
    Gem-2.5-pro     & 21.2 & 24.5 & 19.1 & $-5.4$  & 40.3 \\
    Gem-3-flash     & 21.2 & 23.5 & 19.8 & $-3.7$  & 40.2 \\
    Cl-haiku-4.5    & 21.0 & 24.3 & 19.0 & $-5.3$  & 39.2 \\
    Gem-2.0-fl.lt   & 18.7 & 21.1 & 17.2 & $-3.9$  & 40.1 \\
    Gem-2.5-fl.lt   & 17.1 & 17.6 & 16.8 & $-0.8$  & 36.9 \\
    gpt-4.1         & 16.2 & 18.1 & 15.0 & $-3.1$  & 42.4 \\
    Gem-2.5-flash   & 14.5 & 15.2 & 14.0 & $-1.2$  & 37.5 \\
    o3              & 13.7 & 14.9 & 12.9 & $-2.1$  & 37.6 \\
    o4-mini         & 13.5 & 13.6 & 13.4 & $-0.2$  & 36.0 \\
    o3-mini         & 12.4 & 12.3 & 12.5 & $+0.3$  & 36.5 \\
    o1              & 11.9 & 12.0 & 11.9 & $-0.1$  & 37.7 \\
    gpt-5-mini      & 11.6 & 10.9 & 12.0 & $+1.1$  & 40.3 \\
    Gem-3.1-fl.lt   & 10.7 & 11.7 & 10.1 & $-1.7$  & 39.0 \\
    gpt-5           & 10.7 & 12.0 &  9.9 & $-2.1$  & 37.6 \\
    \midrule
    \textbf{Avg.}   & 27.0 & 28.9 & 26.0 & $-2.9$  & 38.2 \\
    \bottomrule
    \end{tabular}
    \end{minipage}\hfill
    \begin{minipage}[t]{0.48\textwidth}\centering
    \textbf{Short-horizon}\\[2pt]
    \begin{tabular}{l r r r r r}
    \toprule
    \textbf{Model} & \textbf{All} & \textbf{Sta.} & \textbf{Evo.} & $\boldsymbol{\Delta}$ & \textbf{Dist.} \\
    \midrule
    Cl-opus-4.5     & 66.2 & 67.1 & 60.5 & $-6.5$  & 50.7 \\
    Cl-3.7-sonnet   & 61.6 & 63.4 & 49.8 & $-13.6$ & 43.5 \\
    Gem-3-pro       & 56.7 & 57.0 & 55.4 & $-1.6$  & 43.5 \\
    Gem-3.1-pro     & 56.7 & 57.0 & 54.5 & $-2.5$  & 40.9 \\
    Cl-opus-4       & 56.4 & 56.9 & 53.2 & $-3.7$  & 46.5 \\
    Cl-sonnet-4     & 47.8 & 48.8 & 40.8 & $-8.0$  & 46.4 \\
    Cl-3.5-haiku    & 46.4 & 49.0 & 29.6 & $-19.4$ & 52.8 \\
    Cl-sonnet-4.5   & 45.3 & 46.1 & 40.3 & $-5.7$  & 46.3 \\
    Cl-3.5-son.-v1  & 35.7 & 35.9 & 35.0 & $-0.9$  & 36.6 \\
    Gem-2.0-flash   & 30.1 & 30.4 & 27.9 & $-2.5$  & 44.5 \\
    gpt-4o          & 28.4 & 29.2 & 23.2 & $-6.0$  & 59.7 \\
    Cl-haiku-4.5    & 26.0 & 26.9 & 20.2 & $-6.7$  & 50.3 \\
    Gem-2.5-pro     & 25.9 & 27.0 & 18.5 & $-8.5$  & 48.7 \\
    Gem-3-flash     & 24.9 & 26.0 & 17.6 & $-8.4$  & 50.6 \\
    Gem-2.0-fl.lt   & 21.3 & 21.9 & 17.2 & $-4.7$  & 46.8 \\
    Gem-2.5-fl.lt   & 17.5 & 18.1 & 13.7 & $-4.4$  & 50.3 \\
    Gem-2.5-flash   & 16.4 & 17.0 & 12.4 & $-4.5$  & 41.1 \\
    gpt-4.1         & 15.9 & 16.7 & 10.7 & $-6.0$  & 53.9 \\
    o3              & 14.7 & 15.1 & 12.4 & $-2.6$  & 45.1 \\
    o4-mini         & 13.2 & 13.9 &  8.6 & $-5.3$  & 47.0 \\
    gpt-5           & 12.9 & 13.1 & 11.6 & $-1.6$  & 45.4 \\
    Gem-3.1-fl.lt   & 12.8 & 13.8 &  6.0 & $-7.8$  & 49.4 \\
    gpt-5-mini      & 11.1 & 11.7 &  7.7 & $-3.9$  & 52.0 \\
    o1              & 11.1 & 11.5 &  8.2 & $-3.4$  & 51.8 \\
    o3-mini         & 10.3 & 10.8 &  7.3 & $-3.5$  & 49.4 \\
    \midrule
    \textbf{Avg.}   & 30.6 & 31.4 & 25.7 & $-5.7$  & 47.7 \\
    \bottomrule
    \end{tabular}
    \end{minipage}
\end{table*}

\begin{table*}[h]
    \caption{Per-model results: \textbf{Explicit rewrite} and \textbf{Implicit rewrite} conditions (981 items each: 375 static, 606 evolved).}
    \label{tab:explicit_detail}\label{tab:implicit_detail}\vspace{-1mm}
    \centering\scriptsize
    \begin{minipage}[t]{0.48\textwidth}\centering
    \textbf{Explicit rewrite}\\[2pt]
    \begin{tabular}{l r r r r r}
    \toprule
    \textbf{Model} & \textbf{All} & \textbf{Sta.} & \textbf{Evo.} & $\boldsymbol{\Delta}$ & \textbf{Dist.} \\
    \midrule
    Cl-3.5-haiku    & 68.2 & 70.8 & 66.6 & $-4.2$  & 57.6 \\
    Cl-3.7-sonnet   & 64.4 & 63.5 & 65.0 & $+1.5$  & 43.8 \\
    Cl-opus-4.5     & 60.2 & 62.3 & 59.0 & $-3.3$  & 52.1 \\
    Gem-3.1-pro     & 45.9 & 49.1 & 43.9 & $-5.2$  & 46.9 \\
    Gem-3-pro       & 44.0 & 48.5 & 41.3 & $-7.3$  & 45.0 \\
    Cl-opus-4       & 42.6 & 46.4 & 40.3 & $-6.1$  & 53.2 \\
    Cl-3.5-son.-v1  & 37.7 & 39.7 & 37.1 & $-2.6$  & 26.0 \\
    Cl-sonnet-4     & 35.5 & 35.2 & 35.6 & $+0.4$  & 48.6 \\
    Cl-sonnet-4.5   & 35.5 & 35.5 & 35.5 & $+0.0$  & 57.3 \\
    gpt-4o          & 25.3 & 27.5 & 23.9 & $-3.5$  & 59.0 \\
    Gem-2.0-flash   & 25.1 & 26.4 & 24.3 & $-2.1$  & 44.4 \\
    Gem-2.5-pro     & 21.7 & 23.5 & 20.6 & $-2.8$  & 49.1 \\
    Cl-haiku-4.5    & 20.3 & 21.1 & 19.8 & $-1.3$  & 48.0 \\
    Gem-3-flash     & 20.3 & 21.9 & 19.3 & $-2.6$  & 53.5 \\
    Gem-2.0-fl.lt   & 17.8 & 19.7 & 16.7 & $-3.1$  & 53.7 \\
    Gem-2.5-fl.lt   & 17.6 & 16.8 & 18.2 & $+1.4$  & 56.8 \\
    gpt-4.1         & 15.8 & 14.7 & 16.5 & $+1.8$  & 56.5 \\
    Gem-2.5-flash   & 14.8 & 17.1 & 13.4 & $-3.7$  & 51.5 \\
    o4-mini         & 14.8 & 14.1 & 15.2 & $+1.0$  & 47.2 \\
    o3              & 12.9 & 12.8 & 13.0 & $+0.2$  & 45.9 \\
    o3-mini         & 12.4 & 12.0 & 12.7 & $+0.7$  & 48.8 \\
    o1              & 12.1 & 12.8 & 11.7 & $-1.1$  & 47.6 \\
    gpt-5           & 11.0 & 10.1 & 11.6 & $+1.4$  & 47.1 \\
    gpt-5-mini      & 10.8 & 11.7 & 10.2 & $-1.5$  & 47.6 \\
    Gem-3.1-fl.lt   & 10.0 & 11.2 &  9.2 & $-2.0$  & 55.8 \\
    \midrule
    \textbf{Avg.}   & 27.9 & 29.0 & 27.2 & $-1.8$  & 49.7 \\
    \bottomrule
    \end{tabular}
    \end{minipage}\hfill
    \begin{minipage}[t]{0.48\textwidth}\centering
    \textbf{Implicit rewrite}\\[2pt]
    \begin{tabular}{l r r r r r}
    \toprule
    \textbf{Model} & \textbf{All} & \textbf{Sta.} & \textbf{Evo.} & $\boldsymbol{\Delta}$ & \textbf{Dist.} \\
    \midrule
    Cl-opus-4.5     & 57.7 & 62.4 & 54.8 & $-7.6$  & 66.8 \\
    Gem-3.1-pro     & 48.7 & 53.2 & 45.9 & $-7.3$  & 53.4 \\
    Cl-3.7-sonnet   & 47.3 & 45.7 & 48.2 & $+2.5$  & 56.4 \\
    Cl-3.5-haiku    & 45.7 & 46.3 & 45.4 & $-0.9$  & 59.0 \\
    Cl-opus-4       & 43.1 & 46.7 & 40.9 & $-5.7$  & 63.0 \\
    Gem-3-pro       & 39.9 & 45.1 & 36.8 & $-8.3$  & 49.5 \\
    Cl-sonnet-4.5   & 37.9 & 42.4 & 35.1 & $-7.3$  & 63.1 \\
    Cl-sonnet-4     & 37.7 & 41.1 & 35.6 & $-5.4$  & 62.1 \\
    Cl-3.5-son.-v1  & 29.7 & 29.5 & 29.6 & $+0.1$  & 43.4 \\
    Gem-2.0-flash   & 24.6 & 28.5 & 22.1 & $-6.4$  & 56.0 \\
    gpt-4o          & 24.0 & 26.9 & 22.1 & $-4.8$  & 63.7 \\
    Gem-3-flash     & 23.1 & 29.1 & 19.5 & $-9.6$  & 66.9 \\
    Gem-2.5-pro     & 22.7 & 27.5 & 19.8 & $-7.7$  & 58.6 \\
    Cl-haiku-4.5    & 20.9 & 25.3 & 18.2 & $-7.2$  & 60.3 \\
    Gem-2.0-fl.lt   & 19.3 & 22.4 & 17.3 & $-5.1$  & 60.9 \\
    Gem-2.5-flash   & 16.1 & 19.5 & 14.0 & $-5.4$  & 53.0 \\
    o4-mini         & 15.8 & 16.0 & 15.7 & $-0.3$  & 56.7 \\
    Gem-2.5-fl.lt   & 15.5 & 14.9 & 15.8 & $+0.9$  & 57.9 \\
    gpt-4.1         & 14.9 & 15.5 & 14.5 & $-0.9$  & 66.0 \\
    o3              & 14.1 & 14.4 & 13.9 & $-0.5$  & 56.7 \\
    o1              & 13.8 & 13.6 & 13.9 & $+0.3$  & 58.1 \\
    gpt-5           & 13.5 & 12.5 & 14.0 & $+1.5$  & 57.0 \\
    o3-mini         & 12.8 & 13.1 & 12.7 & $-0.4$  & 58.2 \\
    gpt-5-mini      & 12.6 & 13.6 & 12.0 & $-1.6$  & 62.4 \\
    Gem-3.1-fl.lt   &  9.5 & 10.9 &  8.6 & $-2.4$  & 59.6 \\
    \midrule
    \textbf{Avg.}   & 26.4 & 28.6 & 25.0 & $-3.6$  & 58.8 \\
    \bottomrule
    \end{tabular}
    \end{minipage}
\end{table*}

\begin{table*}[h]
    \caption{Per-model results: \textbf{Relaxed-filter} condition (majority $\geq$3/5 filter, 3{,}628 items: 1{,}485 static, 2{,}143 evolved).}
    \label{tab:low_subtlety_detail}\vspace{-1mm}
    \centering\scriptsize
    \begin{tabular}{l r r r r r}
    \toprule
    \textbf{Model} & \textbf{All} & \textbf{Sta.} & \textbf{Evo.} & $\boldsymbol{\Delta}$ & \textbf{Dist.} \\
    \midrule
    Cl-opus-4.5     & 73.7 & 75.6 & 72.5 & $-3.1$  & 55.4 \\
    Cl-3.7-sonnet   & 73.0 & 74.5 & 72.0 & $-2.5$  & 51.0 \\
    Gem-3.1-pro     & 65.4 & 67.3 & 64.1 & $-3.2$  & 48.3 \\
    Gem-3-pro       & 65.1 & 67.5 & 63.5 & $-4.1$  & 48.7 \\
    Cl-opus-4       & 62.8 & 64.8 & 61.5 & $-3.4$  & 54.1 \\
    Cl-3.5-haiku    & 60.3 & 62.6 & 58.7 & $-3.8$  & 58.4 \\
    Cl-sonnet-4     & 54.5 & 55.1 & 54.1 & $-0.9$  & 54.2 \\
    Cl-sonnet-4.5   & 50.3 & 52.5 & 48.9 & $-3.6$  & 59.1 \\
    Cl-3.5-son.-v1  & 44.5 & 43.9 & 44.9 & $+1.0$  & 51.1 \\
    Gem-2.5-pro     & 36.9 & 39.9 & 34.8 & $-5.1$  & 55.3 \\
    Cl-haiku-4.5    & 35.6 & 37.0 & 34.7 & $-2.4$  & 57.6 \\
    gpt-4o          & 35.5 & 37.4 & 34.1 & $-3.3$  & 60.7 \\
    Gem-2.0-flash   & 35.4 & 35.5 & 35.3 & $-0.2$  & 46.8 \\
    Gem-3-flash     & 33.3 & 36.7 & 30.9 & $-5.8$  & 57.2 \\
    Gem-2.0-fl.lt   & 26.1 & 27.1 & 25.4 & $-1.7$  & 53.6 \\
    gpt-4.1         & 25.0 & 25.1 & 25.0 & $-0.2$  & 60.0 \\
    Gem-2.5-fl.lt   & 24.4 & 28.4 & 21.7 & $-6.7$  & 61.3 \\
    Gem-2.5-flash   & 24.3 & 27.7 & 22.0 & $-5.7$  & 52.8 \\
    o3              & 24.0 & 23.4 & 24.4 & $+1.0$  & 51.4 \\
    o4-mini         & 22.5 & 22.5 & 22.4 & $-0.0$  & 49.9 \\
    gpt-5           & 21.5 & 20.3 & 22.4 & $+2.1$  & 51.1 \\
    o1              & 19.7 & 20.8 & 18.9 & $-1.9$  & 53.0 \\
    gpt-5-mini      & 19.7 & 20.3 & 19.2 & $-1.2$  & 54.8 \\
    Gem-3.1-fl.lt   & 19.6 & 21.4 & 18.3 & $-3.1$  & 56.9 \\
    o3-mini         & 18.8 & 20.9 & 17.4 & $-3.5$  & 54.1 \\
    \midrule
    \textbf{Avg.}   & 38.9 & 40.3 & 37.9 & $-2.5$  & 54.3 \\
    \bottomrule
    \end{tabular}
\end{table*}

\end{document}

%% file: math_commands.tex
\usepackage{amsmath,amsfonts,bm}

\def\eqref#1{equation~\ref{#1}}

\def\1{\bm{1}}

\DeclareMathAlphabet{\mathsfit}{\encodingdefault}{\sfdefault}{m}{sl}
\SetMathAlphabet{\mathsfit}{bold}{\encodingdefault}{\sfdefault}{bx}{n}

